\documentclass[10pt,twocolumn,letterpaper]{article}

\usepackage{cvpr}              
\usepackage{makecell}
\usepackage{listings}
\usepackage[ruled,vlined,linesnumbered]{algorithm2e}
\DontPrintSemicolon
\SetKwInOut{Input}{Input}
\SetKwInOut{Output}{Output}
\SetKwInOut{Params}{Params}

\usepackage{listings}
\usepackage{xcolor}

\definecolor{cvprblue}{rgb}{0.21,0.49,0.74}
\usepackage[pagebackref,breaklinks,colorlinks,allcolors=cvprblue]{hyperref}

\def\paperID{20709} 
\def\confName{CVPR}
\def\confYear{2026}

\title{VICoT-Agent: A Vision-Interleaved Chain-of-Thought Framework for Interpretable Multimodal Reasoning and Scalable Remote Sensing Analysis}

\author{
Chujie Wang\textsuperscript{1*} \quad
Zhiyuan Luo\textsuperscript{1*} \quad
Ruiqi Liu\textsuperscript{2} \quad \\
Can Ran\textsuperscript{1} \quad
Shenghua Fan\textsuperscript{1$\dagger$} \quad
Xi Chen\textsuperscript{1$\dagger$} \quad
Chu He\textsuperscript{1$\dagger$} \\
\textsuperscript{1}Wuhan University \quad\textsuperscript{2}University of Toronto
}

\makeatletter
\renewcommand\paragraph{\@startsection{paragraph}{4}{\z@}%
  {1.5ex \@plus 0.2ex \@minus 0.1ex}%
  {-0.6em}%
  {\normalfont\normalsize\bfseries}}
\makeatother
\begin{document}
\maketitle
\begin{figure*}[ht]
\centering
\includegraphics[width=1.0\textwidth]{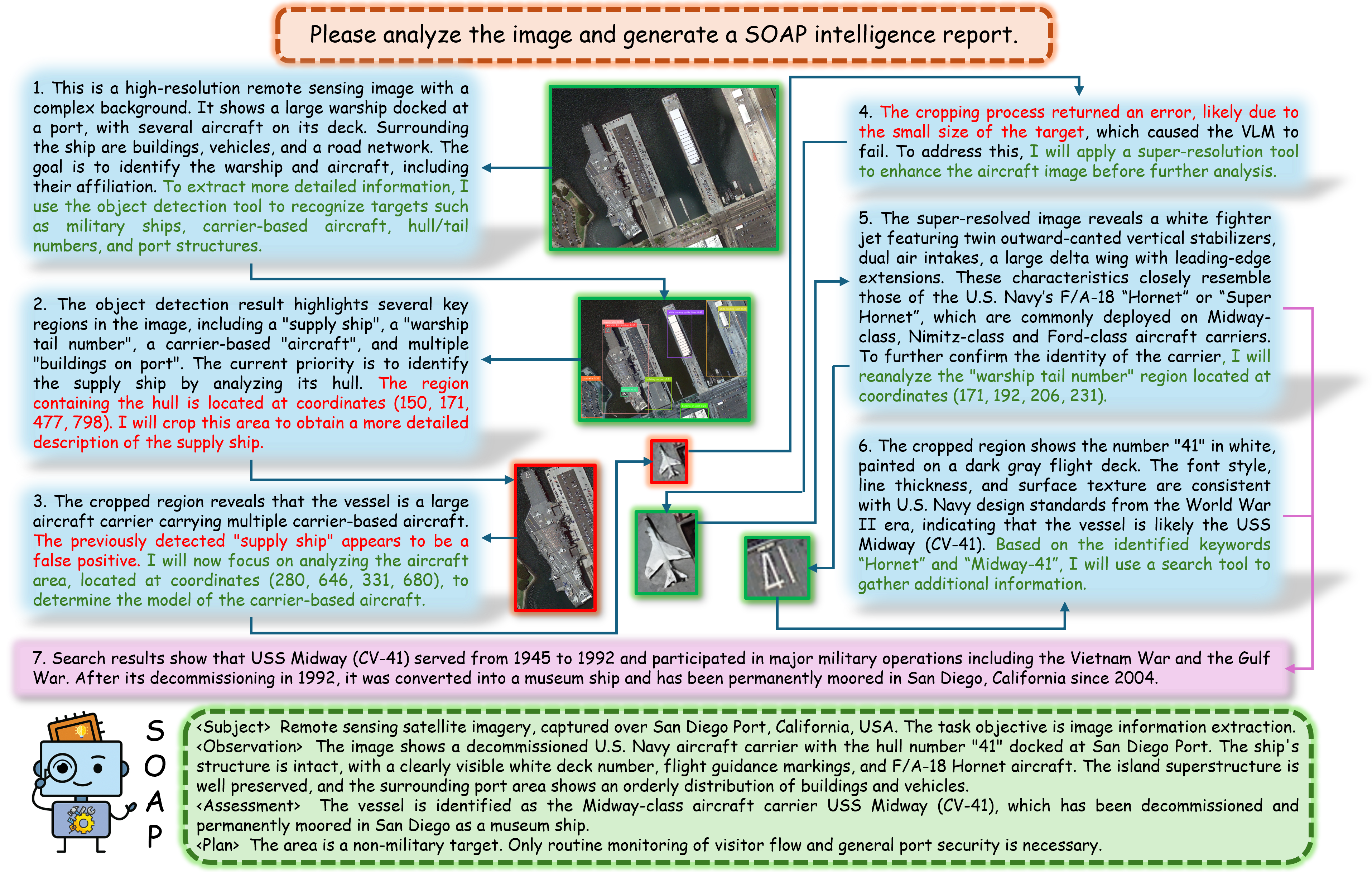} 
\caption{Example of multi-round reasoning and tool invocation by the VICoT Agent in a remote sensing image analysis task. The agent begins by describing the overall scene and identifying relevant objects using an open-vocabulary detector (Steps 1–2). It then applies cropping (Step 3) and super-resolution (Step 4) to refine visual details, followed by augmented aircraft type recognition (Step 5) and hull number identification (Step 6). Based on extracted keywords, it invokes a web search tool to retrieve supporting background information (Step 7), final in a SOAP-formatted intelligence report. Each reasoning step is grounded in a specific visual region and associated tool output, demonstrating the interpretability and transparency.} 
\label{fig1}
\end{figure*}
\begin{abstract}
The current remote sensing image analysis task is increasingly evolving from traditional object recognition to complex intelligence reasoning, which places higher requirements on the model's reasoning ability and the flexibility of tool invocation. To this end, we propose a new multimodal agent framework, Vision-Interleaved Chain-of-Thought Framework (VICoT), which implements explicit multi-round reasoning by dynamically incorporating visual tools into the chain of thought. Through a stack-based reasoning structure and a modular MCP-compatible tool suite, VICoT enables LLMs to efficiently perform multi-round, interleaved vision-language reasoning tasks with strong generalization and flexibility.We also propose the Reasoning Stack distillation method to migrate complex Agent behaviors to small, lightweight models, which ensures the reasoning capability while significantly reducing complexity. Experiments on multiple remote sensing benchmarks demonstrate that VICoT significantly outperforms existing SOTA frameworks in reasoning transparency, execution efficiency, and generation quality.
\end{abstract}

\section{Introduction}
\label{sec:intro}

Recently, analysis on remote sensing images has evolved from traditional object recognition and identification to object interpretation, geo-localization, and intelligence extraction \cite{congeo,geochat,earthdial}. This demand is closely related to the evolving goals of the AI field. Some approaches have adopted multimodal methods, incorporating large-scale remote sensing datasets into pretraining to build remote sensing-specific MLLMs \cite{skysense}. However, remote sensing tasks often require fine-grained multi-scale analysis, typically involving post-processing and multiple rounds of reasoning \cite{earthgpt}. Relying solely on the capabilities of large language models is insufficient to handle the entire complex workflow \cite{toolstar}. Furthermore, models with massive parameter sizes are not ideal for deployment in edge or satellite-based environments, which are common in remote sensing. Agent as an external LLM framework, is well-suited for complex reasoning tasks. Agent can understand roles and environmental constraints, autonomously invoke tools, and offer high automation \cite{agentsurvey,multiagentsurvey}. When integrated with the necessary visual-language tools for remote sensing, agent is ideal for highly autonomous and detailed image analysis tasks. However, most of the main agent frameworks are still limited to predefined workflows with single-turn \cite{langgraph,rsagent}. In image analysis, multimodal agent often functions as executors, returning workflow output (such as image information or RAG results) to blend into the prompt of LLM for summarization. It often produces suboptimal results.

Agent thinking is probably the best way to break free from predefined workflows. The most straightforward approach is to add a planning module after the image input, allowing the agent to generate the execution workflow based on a custom plan dynamically \cite{visiongpt}. To further improve this, the agent incorporates a memory module, enabling the agent to engage in multi-round interactions between (re) planning and execution, refining the workflow step by step until it achieves optimal results \cite{hugginggpt}. However, this method consumes massive tokens and storage space, resulting in significant redundancy. Although it may work for short tasks such as decryption or number problem that require only 1–3 interaction rounds, it struggles to support remote sensing tasks, which often require more than three rounds of visual interaction and are aimed at edge deployment. Moreover, this "split between brain and limbs" design goes against the original intention of using agents to simulate how humans solve complex problems. Since LLMs inherently have reasoning capabilities \cite{cot}, why not embed the multi-step execution directly into the Chain-of-Thought (CoT) process?

To address these challenges, we propose the VICoT Framework (Vision-Interleaved Chain-of-Thought). It integrates reasoning and tool invocation by dynamically inserting and visualizing visual tool output within the chain-of-thought. We design a dynamic reasoning stack to manage and store the reasoning trajectory and also package the tool set using the Model Context Protocol (MCP) protocol \cite{mcp}, enabling one-to-many matching with each stack entry. This design remains modular and plug-and-play by keeping tools isolated from the agent’s runtime environment. VICoT Agent treating workflow construction and execution as a matching process between the stack and the tool set. For remote sensing image analysis, we introduce a range of visual-language tools, including open-vocabulary object detection, cloud/rain removal, super-resolution, RAG and so on. These cover most scenarios encountered in remote sensing applications. To better support ultra-high-resolution (UHR) imagery commonly found in remote sensing, VICoT introduces a hierarchical visual decomposition head that adaptively partitions large-scale inputs into semantic subregions and fuses them through global contextual reasoning. To enhance feasibility and security on edge devices, we propose Reasoning Stack Distillation, using the reasoning stack generated by VICoT-gpt4o to fine-tune a smaller model, Qwen3-14B \cite{qwen3}. After distillation, the student agent VICoT-qwen has a reduced size of only 12 GB (AWQ 4bit) and can run on devices with 16 GB of VRAM, representing a significant drop in inference latency and improving BLEU scores on complex reasoning chains by 3 points.To validate the agent's generalization in remote sensing tasks and capability, we conducted experiments on five remote sensing datasets, among which three are of low to medium resolution and two feature ultra-high-resolution (UHR) imagery. Result shows that VICoT Agent consistently surpasses SOTA RS-MLLMs. Additionally, we quantified performance under existing frameworks and found that, compared to the Plan-Execute architecture, VICoT reduces token consumption by 65\% and latency by 48\% with similar accuracy.

In summary, the main contributions are as follows:
\begin{itemize}
    \item We propose the VICoT stack-based reasoning mechanism and develop the first multimodal agent framework capable of explicit, multi-round vision-language interleaved reasoning. And we prove the framework features modular components that do not primarily rely on built-in model capabilities, offering high autonomy and interpretability.
    \item We introduce the MCP protocol to integrate powerful vision-language tools into a unified and interactive tool set. We ensure that the set remains modular and can easily adapt to different task objectives by isolating the tool environment from the agent framework.
    \item We propose a Reasoning Stack distillation method and construct the first multimodal multi-turn reasoning dataset for remote sensing, VICoT-HRSC, enabling 7B and 14B models to perform multi-turn reasoning and accurate tool invocation without large-scale pretraining, and further validating the effectiveness of this approach across multiple remote sensing datasets.
    \item We develop a highly interpretable Region-Aware Captioned Prompting head for VICoT, which serves as a dedicated UHR image processing module to enhance regional understanding and structured reasoning in large-scale remote sensing scenes.
\end{itemize}


\section{Related Work}
\label{sec:related}

Since ReAct \cite{react} first introduced the static human-like reasoning paradigm of "think-act-observe," enhancements to Chain-of-Thought (CoT) reasoning have diverged into two main directions:

\paragraph{Prompt-driven, code-generation-based LLMs.}

PAL \cite{pal} initially transformed static reasoning patterns into dynamic code-generation processes. MCoT \cite{multimodel} extended reasoning from textual analysis to image understanding. ViperGPT \cite{vipergpt} further encouraged the use of Python for end-to-end multimodal reasoning. Self-Discover \cite{selfdiscover} went beyond dynamic code generation, enabling models to autonomously select from multiple atomic reasoning skills. However, the seemingly dynamic reasoning abilities of these LLMs fundamentally rely on static internal knowledge acquired during pre-training.

\paragraph{Agent-driven, tool-enhanced LLMs.}

Architectures such as Toolformer \cite{toolformer} and ART \cite{art} significantly advanced Function Calling (FC), allowing CoT reasoning to be augmented through external models. However, due to the complexities in tool interaction, some studies abandoned ReAct chain structure. Instead, they adopts a separate ‘plan-execute’ paradigm, in which reasoning and task operations are performed independently and later integrated into a unified output. For example, DDCoT \cite{ddcot} and VoCoT \cite{vocot} rely on predefined algorithms to invoke external visual information, while HuggingGPT \cite{hugginggpt} employs an LLM as a dynamic planning controller, achieving dynamic language-based tool invocation. However, such approaches deviate from the interpretability and chain-like reasoning structure, ultimately falling short of OpenAI’s vision of truly ‘thinking with images’ \cite{think}. DetToolChain \cite{dettoolchain} successfully leveraged LLM-driven CoT to enhance object detection tools, offering valuable insights. This raises an inspiring question: Why not inversely utilize mature and lightweight visual tools to augment CoT reasoning? VisualSketchPad \cite{visualsketchpad} was an initial exploration in this direction but essentially followed a code-generation paradigm, invoking professional drawing tools via code during CoT act steps. More critically, it did not establish a complete agent framework and heavily relied on the inherent visual reasoning capabilities of MLLM.

In summary, there is a pressing need to propose a unified agent framework capable of enhancing CoT with lightweight visual tools, supporting iterative interactions, and dynamically managing multimodal information.

\section{Method}
\label{sec:method}

\begin{figure*}[t]
\centering
\includegraphics[width=1.0\textwidth]{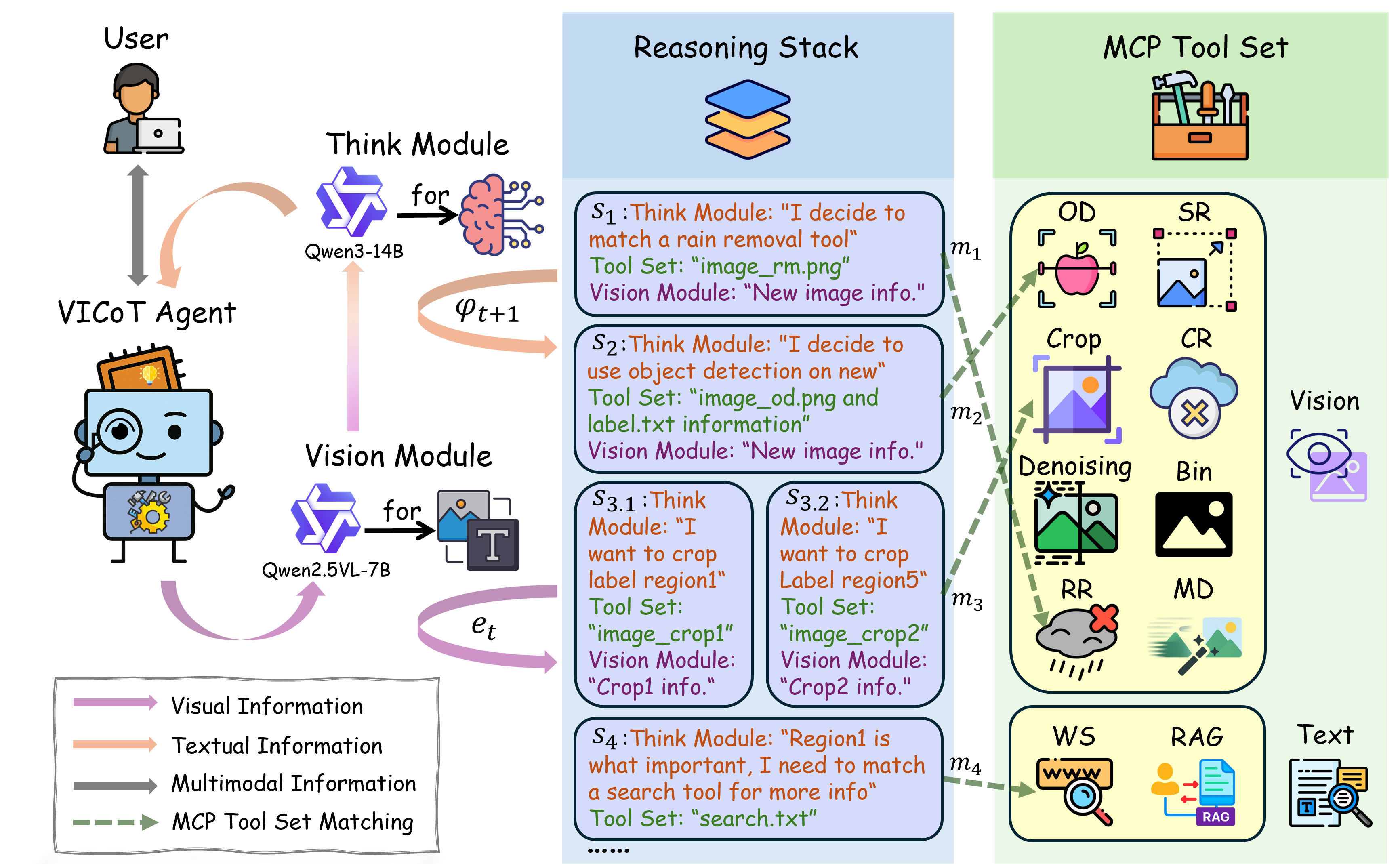} 
\caption{Illustration of the interaction between the Reasoning Stack and the MCP Tool Set in the VICoT framework: The agent operates through a multi-round reasoning process, where each step recorded in the Reasoning Stack(left panel) includes the decision made by the LLM-based Think Module and the corresponding tool invocation. MCP Tool Set(right panel) includes various vision and text tools such as object detection (OD), cropping (Crop), super-resolution (SR), denoising, binarization (Bin), cloud/rain removal (CR/RR), motion deblurring (MD), web search (WS), and RAG-based retrieval. Each tool is invoked via standardized XML under MCP protocol. This design enables modular, interpretable decision-making and flexible tool usage throughout the multi-round reasoning process.}
\label{fig2}
\end{figure*}

\begin{figure*}[htbp]
\centering
\includegraphics[width=0.9\textwidth]{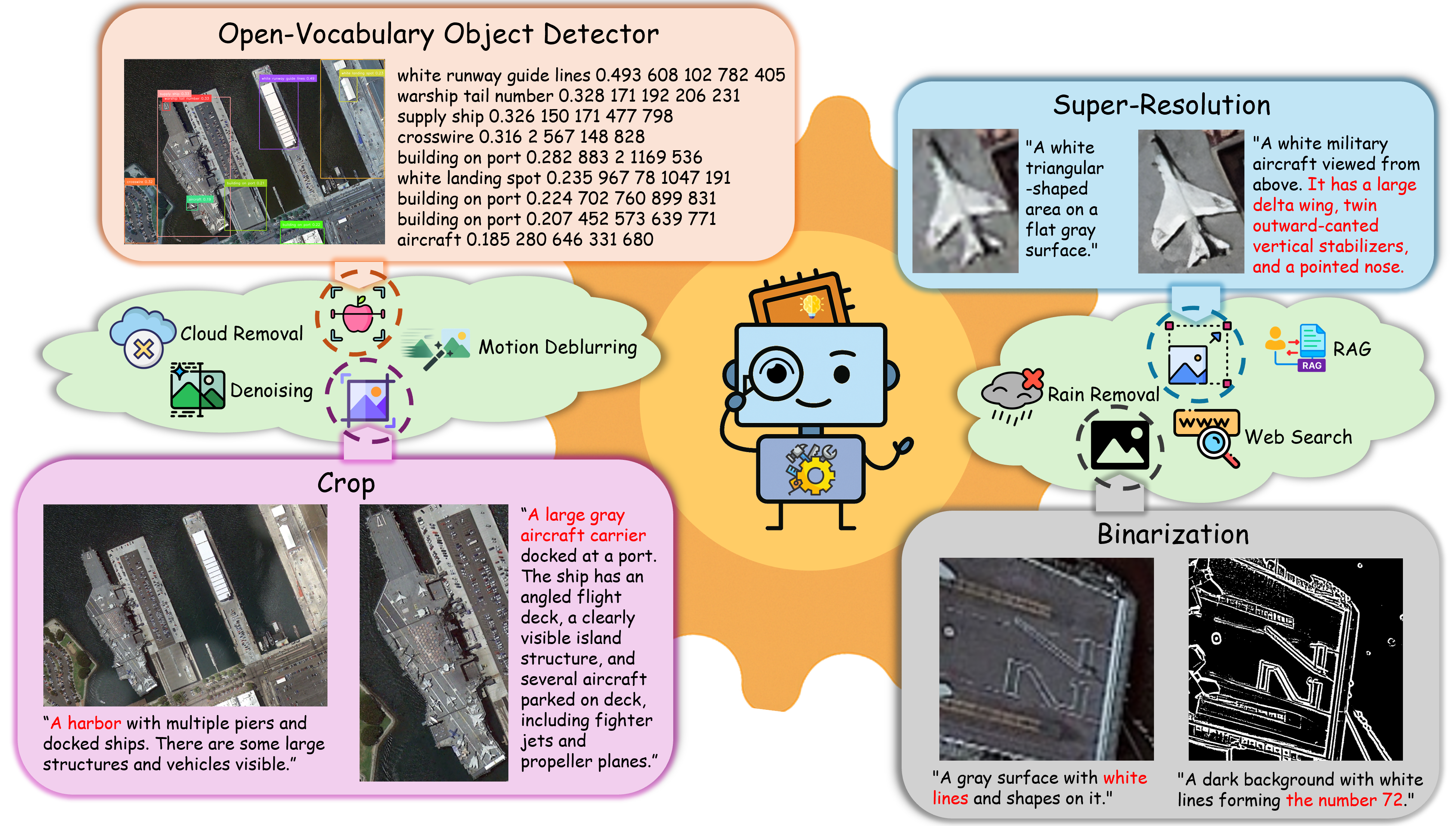} 
\caption{The RS tool-set of the VICoT Agent: The figure shows an output example of the OD tool (top left), as well as the enhancement effects of the Crop, Binarization, and Super-Resolution tools on fine-grained image details.}
\label{fig3}
\end{figure*}

In this section, we systematically present the proposed Agent architecture, elaborating on its design principles and implementation mechanisms, and providing an in-depth analysis of the improvements achieved in key performance metrics.

\subsection{Overview Framework}

VICoT is a multimodal Agent framework that integrates both reasoning and tool invocation, enabling the dynamic insertion and visualization of visual tool processing flows and results within the chain of thought as shown in Figure \ref{fig2}. Our Agent contains two plug-and-play modules: \textbf{Think Module} is generally built upon a reasoning-capable Large Language Model (LLM). The LLM is enhanced through either prompt engineering (for models with strong instruction-following capabilities) or instruction tuning (for lightweight models). The enhancement enables the model to perform contextual reasoning at each turn of the dialogue and decide whether to perform tool invocation. \textbf{Vision Module} constrained to vision-to-text models(VLM) and serves as a bridge between visual tools and the LLM. It converts inputs such as raw images, cropped regions, and enhanced outputs into descriptive text that can be understood by the LLM when processing visual tasks.

These two modules serve as the "neurochip" and "magnifying glass" of the VICoT agent. They manage the interaction between the memory module (Reasoning Stack) and the tool component (MCP Tool Set). This design enables multi-round reasoning and result generation with both visual and textual information in complex tasks, while maintaining high transparency and interpretability.

\subsection{Stack-based Reasoning Modeling}

\paragraph{Why a Stack?}

Imagine a remote sensing analyst continuously tracking an aircraft carrier group in a series of multispectral images: she 'talks to herself,' selects boxes, reviews the results, and then adjusts her line of thinking. In current state-of-the-art agents, each step requires re-feeding the entire history back into the model, causing an exponential increase in tokens and leading to a loss of interpretability in the reasoning chain.

We mirror the analyst’s desk: a stack on which every line of thought, every tool call, every piece of evidence is pushed in order. The top of the stack is all the LLM needs to extend the reasoning; deeper frames remain accessible but don’t inflate the prompt. This structure compresses context, preserves causality, and exposes intermediate evidence—exactly what human analysts do. Therefore, Plan (reasoning) and Execute (tool interaction) are unified into a continuous and automatic process of matching the historical reasoning stack with the tool set.

\paragraph{Formal Definition.}

During the $t$ round of reasoning, the agent maintains a reasoning stack $S_t = [s_1, \dots, s_t]$, and each stack frame is represented as: 
\begin{equation}
s_t = (\varphi_t, m_t, e_t)
\end{equation}
where $\varphi_t$ represents the current decision made after the reasoning process, $m_t = \langle \tau_{i_t}, a_{i_t} \rangle$ (tool+arguments) represents the tool matching process, and $e_t$ represents the returned Evidence (tool output converted to text by the Vision Bridge). For each $s_t$:
\begin{equation}
\varphi_t =
\left\{
  \begin{array}{ll}
    h_\theta(x, Prompt) & \quad t = 1 \\
    h_\theta(S_{t-1})            & \quad t \geq 2
  \end{array}
\right.
\end{equation}
\begin{equation}
m_t = g_\theta(\varphi_t, \mathcal{T})
\end{equation}
\begin{equation}
e_t = \tau_{i_t}(\alpha_{i_t}), \quad S_t = S_{t-1} \parallel (\varphi_t, m_t, e_t)
\end{equation}
When $t=1$, the LLM makes the first decision $\varphi_1$ based on the user’s input $x$ and the system prompt. Subsequently, each decision made by the LLM relies on the current stack state $S_{t-1}$. The function $g_\theta$ vectorizes $\varphi_t$, retrieves $\mathcal{T}$, and selects the highest-scoring match to generate an match $m_t$, which aligns the current stack $S_{t-1}$ with $\mathcal{T}$ and returns the result. The result of this round is processed by $e_t$ and then pushed into the stack, updating the stack state to $S_t$.

When several tools are equally plausible, we duplicate the top frame $s_t$ to spawn a parallel-stack pool of width $\mathcal{W}_t$.
\begin{equation}
\mathcal{P}_t = \left\{ S_t^{(1)}, \dots, S_t^{(W_t)} \right\}
\end{equation}
Each stack proceeds independently for one step; scores are evaluated by a lightweight heuristic from LLM, correct stacks are retained and others discarded.

\paragraph{Complexity Analysis.}

We compared the complexity and efficiency of VICoT and traditional Plan–Replan Loop strategies in remote-sensing reasoning tasks. 

\begin{table}[ht]
\caption{Complexity of VICoT and Plan--Replan Loop.}
\centering
\small
\setlength{\tabcolsep}{4.5pt}
\begin{tabular}{l p{2.7cm} p{2.7cm}}
\toprule
\textbf{Metric} & \centering\textbf{VICoT} & \centering\textbf{Plan$\rightarrow$Replan} \tabularnewline
\midrule
Context tokens & \centering $O(k) \to O(T)$ & \centering $O(t) \to O(T^2)$ \tabularnewline
Tool calls & \centering $\leq 1$ (or $\leq B$) & \centering $\leq m$ scans $+$ 1 exec \tabularnewline
\bottomrule
\end{tabular}
\vspace{2pt}

\label{tab1}
\end{table}


Compared to the Plan–Replan Loop, VICoT reduces the context complexity from quadratic to linear and eliminates full tool scans, achieving both lower token usage and reduced latency. With a sliding window of $k$ frames ($k \ll T$), VICoT reduces context length by up to $(T/2)$-fold and removes $m$-fold tool-scan overhead, which shows $-65\%$ tokens and $-48\%$ latency in experiment. These results demonstrate that VICoT achieves significant improvements in both computational complexity and practical efficiency over conventional methods.

\begin{algorithm}[htbp]
\caption{VICoT Agent Reasoning Workflow}
\label{alg:vicot}
\SetAlgoLined
\KwIn{UHR image $I$, user query $Q$, tool set $\mathcal{T}$}
\KwOut{Structured report $R$}

\BlankLine
Initialize reasoning stack $\mathcal{S} \leftarrow [\,]$\;
Generate initial visual caption $C_0 \leftarrow$ VLM($I, Q, \mathcal{B}$)\;
Push $(I, Q, C_0)$ into $\mathcal{S}$\;

\While{not Terminated}{
  \tcp{Internal Reasoning}
  Generate reasoning trace $T_i \leftarrow$ Think($\mathcal{S}$);

  \tcp{Tool invocation}
  Select tool $t_i \in \mathcal{T}$ based on context of $T_i$;
  Execute $o_i \leftarrow t_i(\text{args from } T_i)$ and receive XML-formatted result;

  \tcp{Observation \& Update}
  Append $\langle T_i, t_i, o_i \rangle$ to $\mathcal{S}$;
  \tcp{Stack-Based State Search}
  \While{$\mathcal{S}$ not empty}{
      $s \leftarrow \mathcal{S}.\text{pop}()$;
      \If{$s$ is discoverable}{
          \textbf{goto} $s$;
      }
  }
  \textbf{break};
}

\BlankLine
\textbf{return} $R \leftarrow$ GenerateSOAPReport($\mathcal{S}$)\;
\end{algorithm}

\subsection{MCP Compatible Tool Suite}

\paragraph{Why use MCP?}

The interaction between the stack and the set is indeed highly efficient, but it requires that the decision $\varphi_t$ of every $S_t$ uses the same Function Calling (FC) method achieving a one-to-many format match with the entire tool-set. Traditional agents often fail to support this retrieval method because the formats used by the model and tool APIs are inconsistent. For example, models like GPT-4o rely on prompts and pretraining to map outputs from various tools to corresponding JSON structures. However, this approach severely limits the number of tools that can be used (usually fewer than 6) \cite{fc} and is highly dependent on the quality of the data used in pretraining for FC. For smaller models that have not undergone similar training, they may even fail to serve as the core of such agents.

To overcome these limitations, we discard the complex and varied FC formats and instead standardize the output of tool parameters in XML format from LLM. We then encapsulate the tools using the Model Context Protocol (MCP) protocol, enabling a single XML output to match and retrieve from the entire encapsulated toolset. 

The tools are further organized into two high-level categories—\textbf{Vision Tools and Text Tools} as shown in Figure \ref{fig2}, allowing the agent to perform coarse-to-fine tool selection: it first chooses the appropriate tool category rather than receiving the full toolset at once, and then selects the specific tool within that category. This approach does not depend on the model size, native capabilities, or the specific tool types, making it more versatile and applicable to various tools. Moreover, the encapsulation of tools and the isolation of their environment from the architecture makes remote calls and tool updates much more convenient.

In our framework, each tool call is constructed as a compact XML block using the tags \texttt{<use\_mcp\_tool>}, \texttt{<server\_name>}, \texttt{<tool\_name>}, and \texttt{<arguments>}. This unified structure enables the agent to generate tool invocations in a consistent format.


\paragraph{RS Tool Set.}



We have integrated 10 different tools for our VICoT Agent as shown in Figure \ref{fig3}: \textbf{The Open-Vocabulary Object Detector tool} \cite{gdino} can perform zero-shot object detection in complex scenarios based on the prompt parameters provided by $\varphi_t$. For remote sensing-specific datasets, we further fine-tuned the tool to improve annotation performance; \textbf{The Crop tool} can extract a single region based on the annotations from the Object Detector tool; \textbf{The Super-Resolution and Binarization tools} \cite{realesrgan} can improve the quality of $e_t$ when Think Module judges the quality of $e_{t-1}$ is unsatisfactory; \textbf{The Cloud Removal, Rain Removal, Denoising, and Motion Deblurring tools} \cite{denoise,restormer} are designed to enhance agent performance under specific conditions; \textbf{The Search and RAG tools} supplement intelligence information via the web and a vector database through keywords from $\varphi_t$. The autonomous method avoids indiscriminate information mixing.

\subsection{Reasoning Stack Distillation}

Although the two modules in the framework are designed to be plug-and-play, experiments show that when the model in the Think Module has a relatively small number of parameters and weak instruction-following ability (e.g., 7B or 14B models), the agent tends to make frequent logical errors. While it still retains the basic capabilities of multimodel tool usage and multi-turn reasoning, the overall performance is noticeably suboptimal.

We proposed a multimodal, multi-turn reasoning dataset named VICoT-HRSC to address the limited adaptability of small models, reduce computational overhead, and minimize token consumption from lengthy prompts—while enabling local deployment and improving edge-side security. This dataset consists of 364 representative images selected from the HRSC \cite{hrsc} dataset, each paired with a corresponding Reasoning Stack generated by prompt-driven GPT-4o within the VICoT framework. Each stack comprises a sequence of multi-turn, stack-structured reasoning steps, including \verb|<think>| reasoning processes, \verb|<tool>| invocations, and the final intelligence outputs.

\begin{figure}[ht]
\centering
\includegraphics[width=0.9\columnwidth]{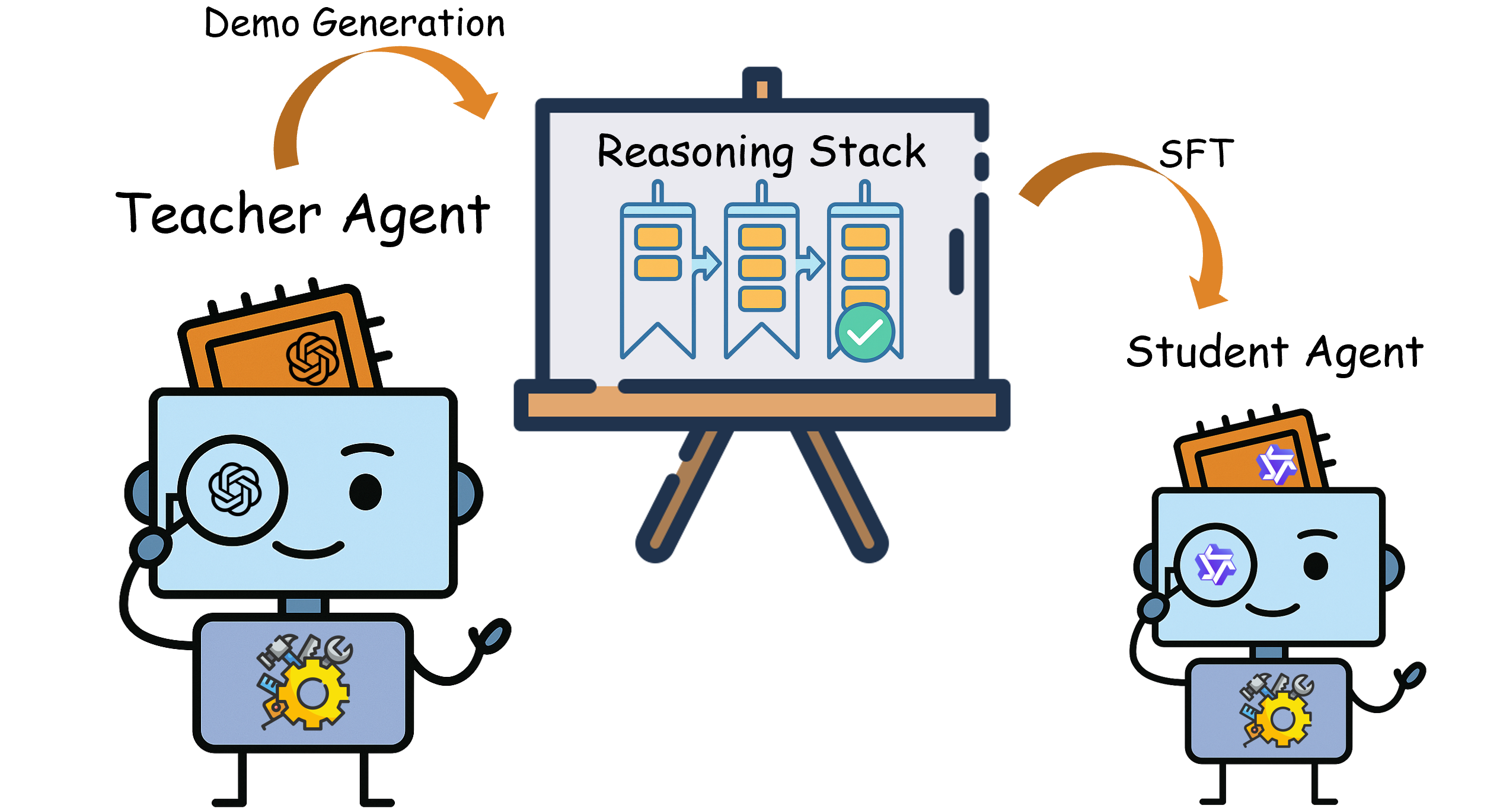} 
\caption{Illustration of Reasoning Stack distillation. The teacher model (left) generates a complete Reasoning Stack, which is used to supervise the student model (right).}
\label{fig4}
\end{figure}

The dataset comprises approximately 4,400 interaction trajectories, with each stack containing an average of 5 to 7 reasoning steps, totaling around 2184K tokens. The dataset supports full reconstruction of the LLM’s reasoning process during training. All tool invocations are represented using standardized XML formats via the MCP interface, ensuring high generality and reusability.

Each trajectory follows a standardized JSON format: (1) a visual description and context initialization, (2) iterative \verb|<think>| reasoning traces with MCP-formatted function call, (3) structured \verb|<tool>| calls with JSON-formatted tool-return results—including intermediate image result paths and corresponding VLM interpretation outputs, and (4) the final SOAP-style intelligence summary; this design enables full interpretability of reasoning flow and modular replay during model training.

We fine-tuned Qwen3-14B on this dataset. After distillation, the model's weights were reduced to just 12 GB with AWQ 4-bit quantization, enabling deployment in a 16 GB GPU memory environment. Under this configuration, each remote sensing image of 1240×980 pixels is processed with an input window of 3–5k tokens and an output window of 0.5–2k tokens, resulting in a full interpretation time of 10–15 seconds per image with stable inference and no memory overflow. This led to a significant reduction in inference latency compared to the teacher agent. Both the dataset and fine-tuning strategy provide critical support for the deployment of VICoT in low-resource edge devices.

\subsection{Region-Aware Captioned Prompting}
To enable effective understanding of ultra-high-resolution (UHR) remote sensing images, we design a region-aware prompting mechanism that integrates spatial decomposition, visual grounding, and multi-turn multimodal reasoning within the VICoT framework.

Given a UHR image and a user instruction, the image is first divided into a grid of fixed-size tiles. Each tile is independently processed by the GroundingDINO model to identify instruction-relevant regions. Tiles without any detected objects are discarded to reduce unnecessary computation and to focus reasoning on meaningful content.

For each valid tile that contains grounded regions, a complete VICoT reasoning cycle is invoked. Instead of a simple captioning process, each tile undergoes the entire intelligent reasoning pipeline, resulting in a coherent region-level language summary that encapsulates both semantic understanding and contextual interpretation.


When all valid tiles have been processed, the system retrieves all stored entries from the Reasoning Stack. Each entry is annotated with its corresponding spatial tag (for example, “Region [2,3]”) to maintain positional alignment. The region-level descriptions are then concatenated sequentially to form a unified contextual narrative, which reflects both semantic content and spatial structure.

Finally, the aggregated regional information is injected into a dedicated UHR Information Integration Prompt. This prompt is specially designed to guide the large language model in synthesizing high-resolution regional insights into a coherent, context-aware response. The LLM processes this structured input holistically and generates the final output based on the combined reasoning traces and the overarching spatial narrative.
\section{Experiments}
\label{sec:exp}

In this section, we focus on four core research questions:
\begin{itemize}
\item Trajectory Quality: Can our framework outperform existing LLM+tool frameworks in trajectory coherence and reasoning quality?
\item Report Quality: Does the VICoT generated report have its advantages compared to real remote sensing intelligence tasks?
\item Performance Driver: What contributes to better performance? — We examine the contributions of the VICoT framework, and distillation strategy. 
\end{itemize}
All reported results are averaged over three runs with different random seeds to ensure statistical stability and reproducibility.

\subsection{Datasets}

The experiment used five datasets: we used our self-built VICoT-HRSC, with 20\% of the data as the validation set to evaluate the accuracy and rationality of tool invocations. RSVQA \cite{rsvqa} includes three types of questions and two resolution versions. RSVQA-LR is the low-resolution version with relatively simple question types, while RSVQA-HR contains high-resolution images and more fine-grained and complex questions. It is aimed at evaluating the agent's performance in answering remote sensing image questions. To evaluate the agent’s performance on ultra-high-resolution (UHR) imagery, we introduce MME-RealWorld-RS \cite{mme} and LRS-VQA-FAIR \cite{lrsvqa}, since FAIR provides high-quality annotations and resolution consistency. All experiments adopt tool settings matched to dataset.

\subsection{Quantitative Evaluation}

\paragraph{Trajectory Quality evaluation.}

We select four representative models \cite{visiongpt,vipergpt,hugginggpt,visualsketchpad}, with GPT-4o as the baseline. We compare them with our proposed VICoT Agent, including both the GPT-4o teacher version and the Qwen3-14B distilled version. Qwen3-14B not distilled version showed high tool-reasoning error and poor convergence, so the baseline was omitted. All models are tested on the VICoT-HRSC validation set to assess their performance in tool invocation tasks. The evaluation includes accuracy, GPT-4.1 rating scores (0-100), human expert assessments (0-100), and BLEU scores (0-1), which reflect tool invocation rationality, trajectory-level reasoning coherence, and overall output quality.


\begin{table}[h]
    \caption{Trajectory quality evaluation on VICoT-HRSC.}
    \label{tab2}
    \centering
    \footnotesize
    \setlength{\tabcolsep}{4.5pt}
    \begin{tabular}{@{}lcccc@{}}
        \toprule
        \textbf{Method} & \textbf{Accuracy} & \textbf{GPT4.1} & \textbf{Humans} & \textbf{BLEU} \\
        \midrule
        VisionGPT           & 61.7   & 59.3   & 60.1   & 0.45 \\
        ViperGPT            & 71.4   & 68.9   & 69.7   & 0.53 \\
        HuggingGPT          & 74.1   & 71.5   & 72.3   & 0.66 \\
        VisualSketchpad     & 73.5   & 70.6   & 71.2   & 0.51 \\
        GPT-4o              & 68.8   & 70.1   & 65.9   & 0.73 \\
        VICoT (4o)         & 92.3~$\uparrow$~23.5 & 91.0~$\uparrow$~20.9 & 95.7~$\uparrow$~29.8 & 0.95~$\uparrow$~0.22 \\
        VICoT (distill)    & 88.6~$\downarrow$~3.7 & 87.4~$\downarrow$~3.6 & 90.5~$\downarrow$~5.2 & 0.98~$\uparrow$~0.03 \\
        \bottomrule
    \end{tabular}
\end{table}


\begin{table*}
    \caption{Comparison of Accuracy on RSVQA-LR and RSVQA-HR datasets.}
    \label{tab3}
    \centering
    \footnotesize
    \setlength{\tabcolsep}{4.5pt}
    \begin{tabular}{@{}lcccc|ccc|cc@{}}
        \toprule
        \makecell{\vspace{-9pt}\textbf{Method}} &
\multicolumn{4}{c|}{\makecell{\textbf{RSVQA-LR}}} &
\multicolumn{3}{c|}{\makecell{\textbf{RSVQA-HR}}} &
\multicolumn{1}{c}{\makecell{\textbf{MME-} \\ \textbf{RW-RS}}} &
\multicolumn{1}{c}{\makecell{\textbf{LRS-} \\ \textbf{FAIR}}} \\
& \makecell{Rural/Urban} 
& \makecell{Presence} 
& \makecell{Compare} 
& \makecell{Average} 
& \makecell{Presence} 
& \makecell{Compare} 
& \makecell{Average} 
&  & \\
        \midrule
        LLaVA-1.5 \cite{llava}         & 59.22\% & 73.16\% & 65.19\% & 65.86\% & 48.96\% & 59.02\% & 53.99\% & 26.38\% & 18.76\% \\
        InstructBLIP \cite{instructblip}      & 62.62\% & 48.83\% & 63.92\% & 59.12\% & 62.63\% & 62.90\% & 62.77\% & -- & -- \\
        mPLUG-Owl2 \cite{mplugowl2}        & 57.99\% & 74.04\% & 65.04\% & 65.69\% & 47.60\% & 58.47\% & 53.04\% & 23.71\% & -- \\
        MiniGPTv2 \cite{minigptv2}         & 60.02\% & 51.64\% & 67.64\% & 59.77\% & 68.34\% & 64.71\% & 66.53\% & 23.33\% & -- \\
        Qwen-VL-Chat \cite{qwenvl}      & 62.00\% & 47.65\% & 54.64\% & 58.73\% & 61.75\% & 65.98\% & 63.97\% & 15.14\% & -- \\
        Qwen2.5-VL \cite{qwen2.5vl}      & 48.00\% & 53.60\% & 75.10\% & 65.80\% & 67.80\% & 73.50\% & 70.65\% & 44.81\% & -- \\
        RSGPT \cite{rsgpt}             & 92.00\% & 90.80\% & 91.80\% & 91.53\% & 90.92\% & 90.02\% & 90.47\% & -- & -- \\
        GeoChat \cite{geochat}           & 92.55\% & 90.71\% & 92.17\% & 91.81\% & 58.45\% & 83.19\% & 70.82\% & 28.62\% & 20.17\% \\
        SkyEyeGPT \cite{skyeyegpt}         & 88.93\% & 88.63\% & 75.00\% & 84.52\% & 80.00\% & 80.13\% & 80.07\% & -- & -- \\
        LHRS-Bot \cite{lhrsbot}          & 89.07\% & 88.51\% & 90.00\% & 89.19\% & --      & --      & -- & -- & --      \\
        RS-Agent \cite{rsagent}          & 97.00\% & 91.07\% & 90.58\% & 90.88\% & --      & --      & -- & -- & --      \\
        LRS-VQA  \cite{lrsvqa}          & -- & -- & -- & -- & -- & -- & -- & 41.89\% & 21.85\% \\
        EarthDial  \cite{earthdial}     & 94.00\% & 92.58\% & 92.75\% & 92.70\% & 58.89\% & 83.11\% & 72.45\% & -- & -- \\
        \midrule
        GPT-4o \cite{gpt4o}            & 92.00\% & 87.32\% & 88.12\% & 89.15\% & 70.00\% & 80.68\% & 75.34\% & 28.92\% & 22.15\% \\
        VI-CoT (4o)       & 97.30\% & 90.72\% & 94.00\% & 94.00\%~$\uparrow$~4.85 & 93.00\% & 91.50\% & 92.25\%~$\uparrow$~16.91 & 32.68\%~$\uparrow$~3.76 & 24.56\%~$\uparrow$~2.41 \\
        VI-CoT (distill)  & 97.20\% & 87.80\% & 89.00\% & 91.33\%~$\downarrow$~2.67 & 92.88\% & 85.30\% & 89.09\%~$\downarrow$~3.61 & 47.40\%~$\uparrow$~14.72 & 27.47\%~$\uparrow$~2.91 \\
        \bottomrule
    \end{tabular}
\end{table*}

As shown in Table \ref{tab2}, VICoT (4o) surpasses all existing frameworks with up to +34.1\% accuracy gain over VisionGPT and +24.6\% over HuggingGPT at least. Compared to baseline, VICoT (4o) surpasses GPT-4o by +34.1\% accuracy, +29.8\% GPT-4.1 score, +45.2\% human score, and +30.1\% BLEU, which shows significant improvements in all metrics. It confirms that the gains come from the framework design rather than LLM itself. The distilled version VICoT (Qwen3-14B) also transfers well, with less than 5\% drop in accuracy and even achieving a +3.2\% BLEU gain, indicating better formatted output than its teacher model.

\paragraph{Remote Sensing Report Quality Evaluation.}

To evaluate the report quality of VICoT Agent in remote sensing scenarios (Report Quality), we conduct experiments on four publicly available remote sensing VQA datasets: RSVQA-LR,RSVQA-HR, MME-RealWorld-RS and LRS-VQA-FAIR. We compare VICoT against a range of well-known models designed for remote sensing tasks. RSVQA-LR and RSVQA-HR include three representative question types: Rural/Urban classification, presence detection, and regional comparison, enabling a comprehensive assessment of the model's understanding and structured reporting capabilities in geospatial contexts. MME-RealWorld-RS and LRS-VQA-FAIR are evaluated through ground-truth (GT) answer matching, where each response is typically a short word-level answer. Both datasets feature ultra-high-resolution imagery, making them suitable for assessing the agent’s performance under fine-grained visual conditions.

As presented in Table \ref{tab3}, VICoT (4o) beats several domain-specific remote sensing agents on low- , high- and ultra-high- resolution datasets. And it surpasses baseline by +4.85\% on RSVQA-LR, +16.91\% on RSVQA-HR, +3.76\% on MME-RW-RS and +2.41\% on LRS-FAIR in average accuracy. This shows that the base language model performs poorly on high-resolution datasets when lacking remote sensing tools. Accuracy of GPT-4o drops by over 13\% from LR to HR, falling behind most domain-specific remote sensing models. On the UHR datasets, GPT-4o exhibits a substantial accuracy drop exceeding 40\%, whereas VICoT shows a smaller degradation, indicating enhanced robustness and adaptability in ultra-high-resolution scenarios.

The distilled version, VICoT (distill), has only a small drop of –2.67\% on RSVQA-LR, –3.21\% on RSVQA-HR,while achieving notable gains of +14.72\% on MME-RW-RS and +2.91\% on LRS-FAIR. This result shows after fine-tuning on the VICoT-HRSC dataset, the distilled model can handle RSVQA tasks well. The observed improvement on UHR datasets may also stem from the Qwen2.5-VL’s inherent strength in UHR visual understanding. Thanks to VICoT framework, it performs especially well on high-resolution questions and reaches nearly SOTA accuracy.

\section{Conclusions}

In this paper, we propose VICoT, a vision-interleaved chain-of-thought agent framework designed for interpretable multi-round and multimodal reasoning in remote sensing image intelligence analysis. It establishes a dynamic matching mechanism between the explicit reasoning stack and the MCP tool set, enabling lightweight visual tools to enhance CoT reasoning. We further introduce a reasoning stack distillation method, enabling small models to inherit powerful reasoning and tool-usage capabilities with reduced computational cost and improved deployment efficiency. Extensive experiments on multiple RS datasets demonstrate that VICoT achieves significant gains in reasoning transparency, trajectory quality, and outperforms existing SOTA works. Future work will explore extending the VICoT framework to open-domain scenarios and improving generalization for lightweight models.
{
    \small
    \bibliographystyle{ieeenat_fullname}
    \bibliography{main}

@String(CVPR= {IEEE Conf. Comput. Vis. Pattern Recog.})

@String(ECCV= {Eur. Conf. Comput. Vis.})

@String(ICLR = {Int. Conf. Learn. Represent.})

@String(CVPR  = {CVPR})

@String(ECCV  = {ECCV})

@String(ICLR  = {ICLR})

@misc{fc,
      title="Function Calling and Other API Updates",
      author="{OpenAI}",
      year=2023,
      howpublished="\url{https://openai.com/blog/function-calling-and-other-api-updates}",
      note="Accessed: 2023-06-28"
}

@misc{think,
      title="Think with images",
      author="{OpenAI}",
      year=2025,
      howpublished="\url{https://openai.com/research/index/release}",
      note="Accessed: 2025-04-16"
}

@misc{ddcot,
      title={DDCoT: Duty-Distinct Chain-of-Thought Prompting for Multimodal Reasoning in Language Models}, 
      author={Ge Zheng and Bin Yang and Jiajin Tang and Hong-Yu Zhou and Sibei Yang},
      year={2023},
      eprint={2310.16436},
      archivePrefix={arXiv},
      primaryClass={cs.CV},
      url={https://arxiv.org/abs/2310.16436}, 
}

@inproceedings{cot,
      author={Wei, Jason and Wang, Xuezhi and Schuurmans, Dale and Bosma, Maarten and ichter, brian and Xia, Fei and Chi, Ed and Le, Quoc V and Zhou, Denny},
      booktitle={Advances in Neural Information Processing Systems},
      editor={S. Koyejo and S. Mohamed and A. Agarwal and D. Belgrave and K. Cho and A. Oh},
      pages={24824--24837},
      publisher={Curran Associates, Inc.},
      title={Chain-of-Thought Prompting Elicits Reasoning in Large Language Models},
      url={https://proceedings.neurips.cc/paper_files/paper/2022/file/9d5609613524ecf4f15af0f7b31abca4-Paper-Conference.pdf},
      volume={35},
      year={2022}
}

@misc{pal,
      title={PAL: Program-aided Language Models}, 
      author={Luyu Gao and Aman Madaan and Shuyan Zhou and Uri Alon and Pengfei Liu and Yiming Yang and Jamie Callan and Graham Neubig},
      year={2023},
      eprint={2211.10435},
      archivePrefix={arXiv},
      primaryClass={cs.CL},
      url={https://arxiv.org/abs/2211.10435}, 
}

@misc{selfdiscover,
      title={Self-Discover: Large Language Models Self-Compose Reasoning Structures}, 
      author={Pei Zhou and Jay Pujara and Xiang Ren and Xinyun Chen and Heng-Tze Cheng and Quoc V. Le and Ed H. Chi and Denny Zhou and Swaroop Mishra and Huaixiu Steven Zheng},
      year={2024},
      eprint={2402.03620},
      archivePrefix={arXiv},
      primaryClass={cs.AI},
      url={https://arxiv.org/abs/2402.03620}, 
}

@inproceedings{vocot,
      title = "{V}o{C}o{T}: Unleashing Visually Grounded Multi-Step Reasoning in Large Multi-Modal Models",
      author = "Li, Zejun  and
            Luo, Ruipu  and
            Zhang, Jiwen  and
            Qiu, Minghui  and
            Huang, Xuanjing  and
            Wei, Zhongyu",
      editor = "Chiruzzo, Luis  and
            Ritter, Alan  and
            Wang, Lu",
      booktitle = "Proceedings of the 2025 Conference of the Nations of the Americas Chapter of the Association for Computational Linguistics: Human Language Technologies (Volume 1: Long Papers)",
      month = apr,
      year = "2025",
      address = "Albuquerque, New Mexico",
      publisher = "Association for Computational Linguistics",
      url = "https://aclanthology.org/2025.naacl-long.192/",
      doi = "10.18653/v1/2025.naacl-long.192",
      pages = "3769--3798",
      ISBN = "979-8-89176-189-6"
}

@misc{agentsurvey,
      title={Agent AI: Surveying the Horizons of Multimodal Interaction}, 
      author={Zane Durante and Qiuyuan Huang and Naoki Wake and Ran Gong and Jae Sung Park and Bidipta Sarkar and Rohan Taori and Yusuke Noda and Demetri Terzopoulos and Yejin Choi and Katsushi Ikeuchi and Hoi Vo and Li Fei-Fei and Jianfeng Gao},
      year={2024},
      eprint={2401.03568},
      archivePrefix={arXiv},
      primaryClass={cs.AI},
      url={https://arxiv.org/abs/2401.03568}, 
}

@inproceedings{multiagentsurvey,
      title     = {Large Language Model Based Multi-agents: A Survey of Progress and Challenges},
      author    = {Guo, Taicheng and Chen, Xiuying and Wang, Yaqi and Chang, Ruidi and Pei, Shichao and Chawla, Nitesh V. and Wiest, Olaf and Zhang, Xiangliang},
      booktitle = {Proceedings of the Thirty-Third International Joint Conference on
                  Artificial Intelligence, {IJCAI-24}},
      publisher = {International Joint Conferences on Artificial Intelligence Organization},
      editor    = {Kate Larson},
      pages     = {8048--8057},
      year      = {2024},
      month     = {8},
      note      = {Survey Track},
      doi       = {10.24963/ijcai.2024/890},
      url       = {https://doi.org/10.24963/ijcai.2024/890},
}

@misc{langgraph,
      author       = {langchain-ai},
      title        = {LangGraph: A framework for building Graph-based Applications using Large Language Models},
      year         = 2025,
      howpublished = {\url{https://github.com/langchain-ai/langgraph}},
      note         = {Accessed: 2025-07-28},
}

@misc{rsagent,
      title={RS-Agent: Automating Remote Sensing Tasks through Intelligent Agent}, 
      author={Wenjia Xu and Zijian Yu and Boyang Mu and Zhiwei Wei and Yuanben Zhang and Guangzuo Li and Mugen Peng},
      year={2025},
      eprint={2406.07089},
      archivePrefix={arXiv},
      primaryClass={cs.CV},
      url={https://arxiv.org/abs/2406.07089}, 
}

@inproceedings{react,
      author  = "Yao, Shunyu and Zhao, Jeffrey and Yu, Dian and Du, Nan and Shafran, Izhak and Narasimhan, Karthik R. and Cao, Yuan",
      year    = 2023,
      title   = "{ReAct: Synergizing Reasoning and Acting in Language Models}",
      booktitle = "Proceedings of the Eleventh International Conference on Learning Representations {(ICLR 2023)}",
      address = "Kigali, Rwanda",
      publisher = "ICLR",
}

@inproceedings{toolformer,
      author = {Schick, Timo and Dwivedi-Yu, Jane and Dessi, Roberto and Raileanu, Roberta and Lomeli, Maria and Hambro, Eric and Zettlemoyer, Luke and Cancedda, Nicola and Scialom, Thomas},
      booktitle = {Advances in Neural Information Processing Systems},
      editor = {A. Oh and T. Naumann and A. Globerson and K. Saenko and M. Hardt and S. Levine},
      pages = {68539--68551},
      publisher = {Curran Associates, Inc.},
      title = {Toolformer: Language Models Can Teach Themselves to Use Tools},
      url = {https://proceedings.neurips.cc/paper_files/paper/2023/file/d842425e4bf79ba039352da0f658a906-Paper-Conference.pdf},
      volume = {36},
      year = {2023}
}

@misc{art,
      title={ART: Automatic multi-step reasoning and tool-use for large language models}, 
      author={Bhargavi Paranjape and Scott Lundberg and Sameer Singh and Hannaneh Hajishirzi and Luke Zettlemoyer and Marco Tulio Ribeiro},
      year={2023},
      eprint={2303.09014},
      archivePrefix={arXiv},
      primaryClass={cs.CL},
      url={https://arxiv.org/abs/2303.09014}, 
}

@misc{multimodel,
      title={Multimodal Chain-of-Thought Reasoning in Language Models}, 
      author={Zhuosheng Zhang and Aston Zhang and Mu Li and Hai Zhao and George Karypis and Alex Smola},
      year={2024},
      eprint={2302.00923},
      archivePrefix={arXiv},
      primaryClass={cs.CL},
      url={https://arxiv.org/abs/2302.00923}, 
}

@misc{mcp,
      author       = {Anthropic},
      title        = {Model Context Protocol},
      year         = 2025,
      howpublished = {\url{https://github.com/modelcontextprotocol}},
      note         = {Accessed: 2025-07-28},
}

@inproceedings{gdino,
      author="Liu, Shilong
      and Zeng, Zhaoyang
      and Ren, Tianhe
      and Li, Feng
      and Zhang, Hao
      and Yang, Jie
      and Jiang, Qing
      and Li, Chunyuan
      and Yang, Jianwei
      and Su, Hang
      and Zhu, Jun
      and Zhang, Lei",
      editor="Leonardis, Ale{\v{s}}
      and Ricci, Elisa
      and Roth, Stefan
      and Russakovsky, Olga
      and Sattler, Torsten
      and Varol, G{\"u}l",
      title="Grounding DINO: Marrying DINO with Grounded Pre-training for Open-Set Object Detection",
      booktitle="Computer Vision -- ECCV 2024",
      year="2025",
      publisher="Springer Nature Switzerland",
      address="Cham",
      pages="38--55",
      isbn="978-3-031-72970-6"
}

@inproceedings{realesrgan,
      author    = {Xintao Wang and Liangbin Xie and Chao Dong and Ying Shan},
      title     = {Real-ESRGAN: Training Real-World Blind Super-Resolution with Pure Synthetic Data},
      booktitle = {International Conference on Computer Vision Workshops (ICCVW)},
      year      = {2021}
}

@article{denoise,
      title={Enhanced cycleGAN network with adaptive dark channel prior for unpaired single-image dehazing},
      author={Xu, Yijun and Zhang, Hanzhi and He, Fuliang and Guo, Jiachi and Wang, Zichen},
      journal={Entropy},
      volume={25},
      number={6},
      pages={856},
      year={2023},
      publisher={MDPI}
}

@inproceedings{restormer,
      title={Restormer: Efficient Transformer for High-Resolution Image Restoration}, 
      author={Syed Waqas Zamir and Aditya Arora and Salman Khan and Munawar Hayat and Fahad Shahbaz Khan and Ming-Hsuan Yang},
      booktitle={CVPR},
      year={2022}
}

@article{qwen2.5vl,
      title={Qwen2.5-VL Technical Report},
      author={Bai, Shuai and Chen, Keqin and Liu, Xuejing and Wang, Jialin and Ge, Wenbin and Song, Sibo and Dang, Kai and Wang, Peng and Wang, Shijie and Tang, Jun and Zhong, Humen and Zhu, Yuanzhi and Yang, Mingkun and Li, Zhaohai and Wan, Jianqiang and Wang, Pengfei and Ding, Wei and Fu, Zheren and Xu, Yiheng and Ye, Jiabo and Zhang, Xi and Xie, Tianbao and Cheng, Zesen and Zhang, Hang and Yang, Zhibo and Xu, Haiyang and Lin, Junyang},
      journal={arXiv preprint arXiv:2502.13923},
      year={2025}
}

@article{qwenvl,
  title={Qwen-VL: A Versatile Vision-Language Model for Understanding, Localization, Text Reading, and Beyond},
  author={Bai, Jinze and Bai, Shuai and Yang, Shusheng and Wang, Shijie and Tan, Sinan and Wang, Peng and Lin, Junyang and Zhou, Chang and Zhou, Jingren},
  journal={arXiv preprint arXiv:2308.12966},
  year={2023}
}

@article{hrsc,
      AUTHOR = {Chen, Weiming and Han, Bing and Yang, Zheng and Gao, Xinbo},
      TITLE = {MSSDet: Multi-Scale Ship-Detection Framework in Optical Remote-Sensing Images and New Benchmark},
      JOURNAL = {Remote Sensing},
      VOLUME = {14},
      YEAR = {2022},
      NUMBER = {21},
      ARTICLE-NUMBER = {5460},
      DOI = {10.3390/rs14215460}
}

@misc{toolstar,
      title={Tool-Star: Empowering LLM-Brained Multi-Tool Reasoner via Reinforcement Learning}, 
      author={Guanting Dong and Yifei Chen and Xiaoxi Li and Jiajie Jin and Hongjin Qian and Yutao Zhu and Hangyu Mao and Guorui Zhou and Zhicheng Dou and Ji-Rong Wen},
      year={2025},
      eprint={2505.16410},
      archivePrefix={arXiv},
      primaryClass={cs.CL},
      url={https://arxiv.org/abs/2505.16410}, 
}

@misc{congeo,
      title={ConGeo: Robust Cross-view Geo-localization across Ground View Variations}, 
      author={Li Mi and Chang Xu and Javiera Castillo-Navarro and Syrielle Montariol and Wen Yang and Antoine Bosselut and Devis Tuia},
      year={2024},
      eprint={2403.13965},
      archivePrefix={arXiv},
      primaryClass={cs.CV},
      url={https://arxiv.org/abs/2403.13965}, 
}

@article{earthgpt,
  	author={Zhang, Wei and Cai, Miaoxin and Zhang, Tong and Zhuang, Yin and Mao, Xuerui},
  	journal={IEEE Transactions on Geoscience and Remote Sensing}, 
  	title={EarthGPT: A Universal Multimodal Large Language Model for Multisensor Image Comprehension in Remote Sensing Domain}, 
  	year={2024},
  	volume={62},
  	number={},
  	pages={1-20},
  	doi={10.1109/TGRS.2024.3409624}
}

@article{geochat,
      title={GeoChat: Grounded Large Vision-Language Model for Remote Sensing},
      author={Kuckreja, Kartik and Danish, Muhammad S. and Naseer, Muzammal and Das, Abhijit and Khan, Salman and Khan, Fahad S.},
      journal={The IEEE/CVF Conference on Computer Vision and Pattern Recognition},
      year={2024}
}

@misc{earthdial,
      title={EarthDial: Turning Multi-sensory Earth Observations to Interactive Dialogues}, 
      author={Sagar Soni and Akshay Dudhane and Hiyam Debary and Mustansar Fiaz and Muhammad Akhtar Munir and Muhammad Sohail Danish and Paolo Fraccaro and Campbell D Watson and Levente J Klein and Fahad Shahbaz Khan and Salman Khan},
      year={2025},
      eprint={2412.15190},
      archivePrefix={arXiv},
      primaryClass={cs.CV},
      url={https://arxiv.org/abs/2412.15190}, 
}

@misc{vipergpt,
      title={ViperGPT: Visual Inference via Python Execution for Reasoning}, 
      author={Dídac Surís and Sachit Menon and Carl Vondrick},
      year={2023},
      eprint={2303.08128},
      archivePrefix={arXiv},
      primaryClass={cs.CV},
      url={https://arxiv.org/abs/2303.08128}, 
}

@inproceedings{visualsketchpad,
      author = {Hu, Yushi and Shi, Weijia and Fu, Xingyu and Roth, Dan and Ostendorf, Mari and Zettlemoyer, Luke and Smith, Noah A and Krishna, Ranjay},
      booktitle = {Advances in Neural Information Processing Systems},
      editor = {A. Globerson and L. Mackey and D. Belgrave and A. Fan and U. Paquet and J. Tomczak and C. Zhang},
      pages = {139348--139379},
      publisher = {Curran Associates, Inc.},
      title = {Visual Sketchpad: Sketching as a Visual Chain of Thought for Multimodal Language Models},
      url = {https://proceedings.neurips.cc/paper_files/paper/2024/file/fb82011040977c7712409fbdb5456647-Paper-Conference.pdf},
      volume = {37},
      year = {2024}
}

@inproceedings{hugginggpt,
      author = {Shen, Yongliang and Song, Kaitao and Tan, Xu and Li, Dongsheng and Lu, Weiming and Zhuang, Yueting},
      booktitle = {Advances in Neural Information Processing Systems},
      editor = {A. Oh and T. Naumann and A. Globerson and K. Saenko and M. Hardt and S. Levine},
      pages = {38154--38180},
      publisher = {Curran Associates, Inc.},
      title = {HuggingGPT: Solving AI Tasks with ChatGPT and its Friends in Hugging Face},
      url = {https://proceedings.neurips.cc/paper_files/paper/2023/file/77c33e6a367922d003ff102ffb92b658-Paper-Conference.pdf},
      volume = {36},
      year = {2023}
}

@inproceedings{skysense,
      author={Guo, Xin and Lao, Jiangwei and Dang, Bo and Zhang, Yingying and Yu, Lei and Ru, Lixiang and Zhong, Liheng and Huang, Ziyuan and Wu, Kang and Hu, Dingxiang and He, Huimei and Wang, Jian and Chen, Jingdong and Yang, Ming and Zhang, Yongjun and Li, Yansheng},
      booktitle={2024 IEEE/CVF Conference on Computer Vision and Pattern Recognition (CVPR)}, 
      title={SkySense: A Multi-Modal Remote Sensing Foundation Model Towards Universal Interpretation for Earth Observation Imagery}, 
      year={2024},
      volume={},
      number={},
      pages={27662-27673},
      doi={10.1109/CVPR52733.2024.02613}
}

@misc{visiongpt,
      title={VisionGPT: Vision-Language Understanding Agent Using Generalized Multimodal Framework}, 
      author={Chris Kelly and Luhui Hu and Bang Yang and Yu Tian and Deshun Yang and Cindy Yang and Zaoshan Huang and Zihao Li and Jiayin Hu and Yuexian Zou},
      year={2024},
      eprint={2403.09027},
      archivePrefix={arXiv},
      primaryClass={cs.CV},
      url={https://arxiv.org/abs/2403.09027}, 
}

@misc{instructblip,
      title={InstructBLIP: Towards General-purpose Vision-Language Models with Instruction Tuning}, 
      author={Wenliang Dai and Junnan Li and Dongxu Li and Anthony Meng Huat Tiong and Junqi Zhao and Weisheng Wang and Boyang Li and Pascale Fung and Steven Hoi},
      year={2023},
      eprint={2305.06500},
      archivePrefix={arXiv},
      primaryClass={cs.CV},
      url={https://arxiv.org/abs/2305.06500}, 
}

@misc{rsgpt,
      title={RSGPT: A Remote Sensing Vision Language Model and Benchmark}, 
      author={Yuan Hu and Jianlong Yuan and Congcong Wen and Xiaonan Lu and Xiang Li},
      year={2023},
      eprint={2307.15266},
      archivePrefix={arXiv},
      primaryClass={cs.CV},
      url={https://arxiv.org/abs/2307.15266}, 
}

@misc{skyeyegpt,
      title={SkyEyeGPT: Unifying Remote Sensing Vision-Language Tasks via Instruction Tuning with Large Language Model}, 
      author={Yang Zhan and Zhitong Xiong and Yuan Yuan},
      year={2024},
      eprint={2401.09712},
      archivePrefix={arXiv},
      primaryClass={cs.CV},
      url={https://arxiv.org/abs/2401.09712}, 
}

@misc{mplugowl2,
      title={mPLUG-Owl2: Revolutionizing Multi-modal Large Language Model with Modality Collaboration}, 
      author={Qinghao Ye and Haiyang Xu and Jiabo Ye and Ming Yan and Anwen Hu and Haowei Liu and Qi Qian and Ji Zhang and Fei Huang and Jingren Zhou},
      year={2023},
      eprint={2311.04257},
      archivePrefix={arXiv},
      primaryClass={cs.CL},
      url={https://arxiv.org/abs/2311.04257}, 
}

@misc{minigptv2,
      title={MiniGPT-v2: large language model as a unified interface for vision-language multi-task learning}, 
      author={Jun Chen and Deyao Zhu and Xiaoqian Shen and Xiang Li and Zechun Liu and Pengchuan Zhang and Raghuraman Krishnamoorthi and Vikas Chandra and Yunyang Xiong and Mohamed Elhoseiny},
      year={2023},
      eprint={2310.09478},
      archivePrefix={arXiv},
      primaryClass={cs.CV},
      url={https://arxiv.org/abs/2310.09478}, 
}

@misc{lhrsbot,
      title={LHRS-Bot: Empowering Remote Sensing with VGI-Enhanced Large Multimodal Language Model}, 
      author={Dilxat Muhtar and Zhenshi Li and Feng Gu and Xueliang Zhang and Pengfeng Xiao},
      year={2024},
      eprint={2402.02544},
      archivePrefix={arXiv},
      primaryClass={cs.CV},
      url={https://arxiv.org/abs/2402.02544}, 
}

@article{rsvqa,
      author={Lobry, Sylvain and Marcos, Diego and Murray, Jesse and Tuia, Devis},
      journal={IEEE Transactions on Geoscience and Remote Sensing}, 
      title={RSVQA: Visual Question Answering for Remote Sensing Data}, 
      year={2020},
      volume={58},
      number={12},
      pages={8555-8566},
      doi={10.1109/TGRS.2020.2988782}
}

@article{lrsvqa,
    title={When Large Vision-Language Model Meets Large Remote Sensing Imagery: Coarse-to-Fine Text-Guided Token Pruning},
    author={Luo, Junwei and Zhang, Yingying and Yang, Xue and Wu, Kang and Zhu, Qi and Liang, Lei and Chen, Jingdong and Li, Yansheng},
    journal={arXiv preprint arXiv:2503.07588},
    year={2025}
}

@article{mme,
      title={MME-RealWorld: Could Your Multimodal LLM Challenge High-Resolution Real-World Scenarios that are Difficult for Humans?},
      author={Zhang, Yi-Fan and Zhang, Huanyu and Tian, Haochen and Fu, Chaoyou and Zhang, Shuangqing and Wu, Junfei and Li, Feng and Wang, Kun and Wen, Qingsong and Zhang, Zhang and others},
      journal={arXiv preprint arXiv:2408.13257},
      year={2024}
}

@misc{gpt4o,
      title={GPT-4o System Card}, 
      author={OpenAI and Aaron Hurst and Adam Lerer and Adam P. Goucher and Adam Perelman and Aditya Ramesh and Aidan Clark and AJ Ostrow and Akila Welihinda},
      year={2024},
      eprint={2410.21276},
      archivePrefix={arXiv},
      primaryClass={cs.CL},
      url={https://arxiv.org/abs/2410.21276}, 
}

@misc{llava,
      title={Improved Baselines with Visual Instruction Tuning}, 
      author={Haotian Liu and Chunyuan Li and Yuheng Li and Yong Jae Lee},
      year={2024},
      eprint={2310.03744},
      archivePrefix={arXiv},
      primaryClass={cs.CV},
      url={https://arxiv.org/abs/2310.03744}, 
}

@misc{qwen3,
      title={Qwen3 Technical Report}, 
      author={An Yang and Anfeng Li and Baosong Yang and Beichen Zhang and Binyuan Hui},
      year={2025},
      eprint={2505.09388},
      archivePrefix={arXiv},
      primaryClass={cs.CL},
      url={https://arxiv.org/abs/2505.09388}, 
}

@misc{dettoolchain,
      title={DetToolChain: A New Prompting Paradigm to Unleash Detection Ability of MLLM}, 
      author={Yixuan Wu and Yizhou Wang and Shixiang Tang and Wenhao Wu and Tong He and Wanli Ouyang and Philip Torr and Jian Wu},
      year={2024},
      eprint={2403.12488},
      archivePrefix={arXiv},
      primaryClass={cs.CV},
      url={https://arxiv.org/abs/2403.12488}, 
}
}

\clearpage
\setcounter{page}{1}
\maketitlesupplementary

\section{Appendix}
\label{sec:rationale}
\noindent
This appendix compiles the supplementary materials of VICoT, encompassing the system's reasoning pipeline, prompt specifications, tool-matching mechanisms, dataset formatting rules, and extended experimental observations. Beyond the main paper, these materials clarify how VICoT structures its multi-stage reasoning process, organizes its XML-based tool interfaces, and maintains isolation in its runtime environment. We also provide the detailed format of the VICoT-HRSC dataset alongside additional quantitative and qualitative results, as well as representative error cases that offer deeper insight into the system's behavior and failure modes.

\subsection{Reasoning Stack Walkthrough}
This section illustrates the step-by-step ``stack push'' process of the VICoT agent's multi-turn reasoning. For each turn, we show how the tuple $\phi$ (decision), $m$ (tool invocation), and $e$ (evidence) are generated and pushed onto the reasoning stack, along with the interactions between tool calls and VLM outputs.

\begin{lstlisting}[basicstyle=\ttfamily\footnotesize,breaklines=true,caption={Reasoning walkthrough}]

====VLM First-Round Rough Description====
This is a high-resolution remote-sensing image showing a coastal port area. A large vessel is berthed at the dock, with multiple aircraft lined up on its deck. The surrounding area is neatly organized, featuring several large buildings, densely parked vehicles, and a network of roads-clearly indicating urban development and distinct port functional zones.

====LLM Invoking OVOD Based on VLM Output====
<think>
The image shows a large vessel docked at a port, with several aircraft on its deck and surrounding buildings, vehicles, and road networks. My goal is to determine the identities and affiliations of the ship and aircraft. Therefore, I will call the GroundingDINO object-detection tool to recognize potential targets such as military warships, carrier aircraft, ship/aircraft tail numbers, and port buildings to extract key regions for further analysis and identification.
</think>
<use_mcp_tool>
    <server_name>mcp_vision_server</server_name>
    <tool_name>image_detection</tool_name>
    <arguments>
    {
        "image_path": "test.png",
        "txt_prompt": "military warship . cargo ship . aircraft . warship tail number . aircraft tail number . marking on tail of ship . marking of aircraft . building on port ."
    }
    </arguments>
</use_mcp_tool>

====GroundingDINO + VLM Tool Results Return====
"annotated_image_path": "test_annotated.png",
"boxes": [
    "white runway guide lines 0.348 608 103 782 404",
    "cargo ship 0.332 149 172 477 796",
    "crosswire 0.298 3 567 149 827",
    "building on port 0.297 885 3 1169 533",
    "little white shape 0.294 383 736 413 773",
    "building on port 0.223 703 760 899 831",
    "white helicopter landing circles 0.218 406 705 455 757",
    "warship tail number 0.214 172 192 205 231",
    "aircraft 0.202 280 645 332 680",
    "warship tail number 0.201 337 145 377 177"
],
"vlm_response": "The image shows a port area captured by remote sensing, containing multiple vessels and port facilities. After processing with GroundingDINO, several target regions were detected and annotated as follows:

1. In the upper-left area, a box labeled "cargo ship" with confidence 0.332 at (149, 172, 477, 796). This identifies the main large ship on the port's left side. Although it appears as a cargo ship, it may actually be a military display vessel.

2. Two "warship tail number" detections within the same ship area:
   - The first at (172, 192, 205, 231), confidence 0.214, near the helicopter landing zone, likely marking its hull number.
   - The second at (337, 145, 377, 177), confidence 0.201, near the island superstructure, part of the vessel's identification.

3. A box labeled "aircraft," confidence 0.202 at (280, 645, 332, 680), identifying a carrier-based aircraft parked at the stern, with wings and tail structure visible.

4. A box labeled "white helicopter landing circles," confidence 0.218 at (406, 705, 455, 757), marking the helicopter landing pad with white circular guide symbols.

5. A "little white shape" box, confidence 0.294 at (383, 736, 413, 773), possibly a small deck fixture or ground marking.

6. A "crosswire" box, confidence 0.298 at (3, 567, 149, 827), covering berth infrastructure or line markings-actual category unclear.

7. A "white runway guide lines" box, confidence 0.348 at (608, 103, 782, 404), showing deck guide lines likely for landing assistance.

8. A "building on port" box, confidence 0.297 at (885, 3, 1169, 533), marking a large portside building with lanes and facilities.

9. Another "building on port," confidence 0.223 at (703, 760, 899, 831), near the water's edge with adjacent roads and vegetation.

Overall, the tool accurately detected vessels, deck guide markings, aircraft, helicopter landing zones, and port buildings, reflecting clear port activity and infrastructure. Some labels like "cargo ship" may be misclassifications, with the vessel more fitting a military display ship."

====LLM Analyzes and Invokes Binarization====
<think>
The object detection highlighted several key areas: a suspected warship (misclassified as "cargo ship"), two "warship tail number" regions, one "aircraft," the "white helicopter landing circles," and two port buildings. Two hull-number boxes were found at (172, 192, 205, 231) and (337, 145, 377, 177), which can identify the vessel class and affiliation.
My immediate priority is to recognize the hull number to determine the ship's identity. I will crop the first number region and apply binarization to enhance the number's readability.
</think>
<use_mcp_tool>
    <server_name>mcp_vision_server</server_name>
    <tool_name>image_binary</tool_name>
    <arguments>
    {
        "image_path": "test.png",
        "x1": 172,
        "y1": 192,
        "x2": 205,
        "y2": 231
    }
    </arguments>
</use_mcp_tool>

====Binarization + VLM Tool Results Return====
"cropped_image_path": [
    "test_binary.png"
],
"vlm_response": [
    "This is a zoomed-in detail from the remote-sensing image, showing a group of white digits "41" on a dark-gray background that resembles flight-deck or hull plating. The "4" has a clear slanted vertical stroke connected to a horizontal bar and another vertical on the right, forming a standard "4." The "1" is a single thick vertical line. The bold strokes and sharp edges suggest a flight-deck number marking on an aircraft carrier, matching US Nimitz- or Ford-class deck paint schemes."
]

====LLM Analyzes and Invokes Full-Ship Crop====
<think>
The cropped region shows the white number "41" on a dark deck. The font style, stroke thickness, and deck texture match US carrier deck markings, strongly indicating a specific ship identity.
In US carrier numbering, "41" corresponds to USS Theodore Roosevelt (CVN-71), typically painted between the stern and front island.
To further verify the platform type, I will crop the entire vessel area to analyze hull structure and aircraft deployment patterns.
</think>
<use_mcp_tool>
    <server_name>mcp_vision_server</server_name>
    <tool_name>image_crop</tool_name>
    <arguments>
    {
        "image_path": "test.png",
        "x1": 149,
        "y1": 172,
        "x2": 477,
        "y2": 796
    }
    </arguments>
</use_mcp_tool>

====Full-Ship Crop + VLM Tool Results Return====
"cropped_image_path": [
    "test_cropped.png"
],
"vlm_response": [
    "This is a high-resolution color remote-sensing image showing a US Navy aircraft carrier docked at a port berth. The vessel fills nearly the entire frame, with hull structure, deck layout, embarked equipment, and surrounding environment clearly visible.

The carrier's deck extends vertically from top to bottom. The front-left deck bears the number "41," confirming this as USS Midway (CV-41). Multiple fixed-wing carrier aircraft-such as F/A-18 Hornets and A-4 Skyhawks-are neatly displayed on the deck in exhibition posture, wings spread and tail fins visible.

The island superstructure sits in the upper right, equipped with radar antennas, the bridge, and communication arrays, topped by a dome radar housing. Orange circle markings north of the island suggest helicopter landing or exhibition zones, matching a museum display layout.

Alongside the port, the carrier is neatly berthed with boarding ramps and gangways connecting deck and pier. Dozens of vehicles are orderly parked in lot areas, and adjacent buildings resemble industrial warehouses.

The surrounding water is calm, indicating a harbor or inland bay. The lower-right shoreline curves naturally, with vegetation meeting water. Uniform lighting and clear vessel details suggest its status as a museum exhibit, consistent with the USS Midway Museum in San Diego."
]

====LLM Generates Final SOAP Report====
<think>
The cropped full-ship view reveals a clear flight deck, displayed aircraft, island radar structure, and pier facilities-confirming this is USS Midway (CV-41), now a museum vessel in San Diego. Aircraft are exhibited in static formation, the island superstructure intact, and port parking and visitor ramps evident-indicating a museum status.
No signs of combat deployment are present. Combined with deck number and hull features, this image depicts the USS Midway Museum's permanent display, of historical and cultural significance.
Information is sufficient; now generating the SOAP intelligence report-task complete.
</think>
<S>
Source: 2024 remote-sensing satellite imagery of San Diego Bay, USA; 
Task background: identify potential military platforms and formations in a naval port.
</S>
<O>
The image was captured over San Diego Bay. The primary vessel is the decommissioned aircraft carrier USS Midway (CV-41). High resolution reveals complete hull structure, white deck marking "41," multiple carrier aircraft (F/A-18, A-4), and flight-guidance symbols. The island superstructure remains intact; port infrastructure and vehicle distribution are orderly.
</O>
<A>
Identified as the Midway-class carrier USS Midway (CV-41), now permanently exhibited at San Diego Bay. Deck layout matches US carrier designs; embarked aircraft are in static display, with no combat configuration. Threat level is low; the vessel has no military function, serving solely for commemoration and public education.
</A>
<P>
Given its museum display status, no further military surveillance is required. Recommend classifying this area as a "non-military monitoring zone," with only low-frequency monitoring for civilian traffic and port security. No air or electronic counter-measures are necessary.
</P>

\end{lstlisting}

\subsection{Prompt Details}

\paragraph{Note}
The prompts presented in this section are formatted for readability. 
Before practical use, each prompt must be programmatically formatted to match the VICoT system's input specification. 
The versions shown here may not fully reflect the latest in-system prompts, as minor updates or structural adjustments may have been made during the development cycle.

\begin{lstlisting}[basicstyle=\ttfamily\footnotesize,breaklines=true,caption={Prompt for LLM},label={llm_prompt}]
SYSTEM_PROMPT_CHAT="""
## Task Description:
    The current task is to construct a dataset for a Visual-Language Model (VLM) and a Large-Language Model (LLM). The detailed workflow is as follows:
    1. Input a remote-sensing satellite image; the VLM provides a coarse overall description. Note that in the first round the VLM description must not contain fine-grained, uniquely identifying information.
    2. Feed the overall description to the LLM for analysis. The LLM outputs its reasoning and, based on the available tools, returns the parameters for a tool invocation.
    3. The tool executes and returns its results, which are then described in detail by the VLM.
    4. Feed the detailed description back to the LLM for further analysis. If the information is insufficient, go back to Step 2; if it is sufficient, the LLM integrates the information and, strictly following the SOAP format, outputs the analysis result, ending the task.
        S.O.A.P format
        S (Subject - intelligence background)
        Source (satellite / reconnaissance aircraft / human), time range, mission context.
        O (Objective - objective data)
        Sensor type (optical / SAR / IR), resolution, raw-data characteristics.
        A (Assessment - analysis and evaluation)
        Target identification (spectral match / geometric features), threat level (high / medium / low), relationship prediction, intent inference.
        P (Plan - action plan)
        Recommended measures (strike / jam / continuous monitoring), priority, cooperating units.
    You now play the role of the LLM to build the dataset. The programme will automatically feed you the VLM descriptions of the images; your main task is to analyse the current results.
## Notes:
    1. Do not output any irrelevant content; our dialogue will be stored as the dataset in its entirety.
    2. Do not use code blocks and do not output acknowledgements such as "received".
    3. Do not give next-step suggestions like a chat bot; focus on the image description only.
    4. The task may involve multiple rounds of dialogue; please strictly follow these requirements in every round.
    5. When the LLM outputs, first wrap your reasoning in <think></think>. If you need to call a tool, strictly follow the tool example.
## Tool Usage Example:
    <use_mcp_tool>
        <server_name>mcp-vision-agent</server_name>
        <tool_name>object_detection</tool_name>
        <arguments>
        {
        "image": "image.jpg", 
        "txt_prompt": "military warship"
        }
        </arguments>
    </use_mcp_tool>
## Tool Usage Rules
    The following rules must always be followed when you invoke tools:
    1. Always supply the correct parameters for tools; never use variable names as parameter values-use actual values.
    2. Invoke a tool only when needed: if no extra information is required, do not call a tool; try to solve the task yourself.
    3. If no tool is required, answer directly.
    4. Do not invoke the same tool again with exactly the same parameters.
    5. When using a tool, you must use the XML format in the example above; do not use any other format.
## Available Tools
{available_tools}
## Flow Example
1. First-round VLM coarse description (used as input for Step 2)  
    This is a high-resolution satellite image showing an aircraft carrier berthed at a harbour. Numerous carrier-borne aircraft are neatly parked on deck. A white hull number is present at the stern, but it is too blurred to identify specific digits. Roads, buildings, and vehicles are laid out in an orderly pattern around the port area.
2. Based on the VLM description, the LLM in Round 1 calls the GroundingDINO tool  
    <think> I need to generate ship-and-harbour intelligence from the description. The image shows an aircraft carrier in port, its flight deck packed with aircraft, and a blurred hull number at the stern. Current information is insufficient to confirm the ship's identity or affiliation. I will call the object-detection tool to locate the carrier structure, number areas, and port buildings for later number extraction and intelligence analysis. </think>  
    <use_mcp_tool>  
    <server_name>mcp-vision-agent</server_name>  
    <tool_name>object_detection</tool_name>  
    <arguments>  
    {  
    "image": "original_satellite_image",  
    "txt_prompt": "military warship . cargo ship . aircraft . warship tail number . aircraft tail number . marking on tail of ship . marking of aircraft . building on port ."  
    }  
    </arguments>  
    </use_mcp_tool>
3. Second-round GroundingDINO results  
    <image>bounding-box image produced by GroundingDINO</image>  
    cargo ship 0.393 3 162 1245 653  
    military warship cargo ship 0.379 129 67 1042 409  
    building on port 0.315 357 512 527 649  
    marking on tail of ship 0.276 972 342 1030 388  
    warship tail number aircraft tail number 0.274 270 369 315 410  
    building on port 0.263 73 0 151 46
4. Second-round VLM description of GroundingDINO bounding boxes  
    This is a high-resolution satellite image of an aircraft carrier docked at a harbour. Multiple coloured boxes highlight key regions and their categories:  

    - Upper-left water area: a box labelled "cargo ship", confidence 0.393, coords (3, 162, 1245, 653), covering the main subject; likely a mis-classification.  

    - Main ship area: a large box labelled "military warship cargo ship", confidence 0.379, coords (129, 67, 1042, 409), spanning the entire carrier.  

    - Mid-port: a box labelled "building on port", confidence 0.315, coords (357, 512, 527, 649).  

    - Stern area: a box labelled "marking on tail of ship", confidence 0.276, coords (972, 342, 1030, 388), suspected hull-number region.  

    - Island-foredeck: a box labelled "warship tail number aircraft tail number", confidence 0.274, coords (270, 369, 315, 410), possibly aircraft tail numbers.  

    - Upper-left near-shore: a small box labelled "building on port", confidence 0.263, coords (73, 0, 151, 46).
5. Second-round LLM analysis of the VLM description and binarisation tool call  
    <think> From the detection results I see these key regions:  
    - "military warship cargo ship" (129, 67, 1042, 409): entire carrier, crucial for structural and platform-level ID;  
    - "warship tail number aircraft tail number" (270, 369, 315, 410): aircraft tails on deck, possible numbers;  
    - "marking on tail of ship" (972, 342, 1030, 388): hull-number area at the stern;  
    - "building on port" regions: aid in judging port characteristics;  
    - "cargo ship" region: low confidence, likely a false positive, ignore.  

    Priority is to identify the carrier's overall configuration, so I will crop the "military warship cargo ship" region and perform binarisation to aid later ship-class confirmation. </think>  
    <use_mcp_tool>  
    <server_name>mcp-vision-agent</server_name>  
    <tool_name>crop_and_enhance</tool_name>  
    <arguments>  
    {  
    "image": "original_satellite_image",  
    "crop_box": [129, 67, 1042, 409],  
    "binarize": true  
    }  
    </arguments>  
    </use_mcp_tool>
6. Third-round VLM description of the binarised image  
    This binarised image shows an aircraft carrier docked at port. Hull lines are crisp, the flight-deck outline is regular, a central runway runs bow to stern, and the stern marking zone is clear. Fighter silhouettes are densely packed, with twin tails and swept wings, and the island sits mid-left. Overall deck layout and airframe shapes match those of a U.S. Nimitz-class carrier.
7. Third-round LLM analysis and further tool call  
    <think> The binarised carrier image shows clear structure and deck layout consistent with a U.S. Nimitz-class. Next I need to crop the aircraft-tail region to confirm aircraft type and consistency with the platform. </think>  
    <use_mcp_tool>  
    <server_name>mcp-vision-agent</server_name>  
    <tool_name>crop</tool_name>  
    <arguments>  
    {  
    "image": "original_satellite_image",  
    "crop_box": [270, 369, 315, 410]  
    }  
    </arguments>  
    </use_mcp_tool>
8. Final SOAP output  
    <think> Full intelligence obtained: a satellite image shot mid-May 2025 shows hull number "72" on an aircraft carrier in port. About 30 carrier-borne fighters are neatly parked; the island and flight deck are clear. The number and aircraft type confirm the ship is the U.S. Nimitz-class USS *Abraham Lincoln* (CVN-72), home-ported at Naval Base San Diego. The carrier is berthed with no signs of deployment. Task complete, generating detailed SOAP report. </think>  
    <S> Source: high-resolution satellite imagery of a naval base, May 2025. </S>  
    <O> Hull "72" carrier berthed at San Diego; ~30 fighters on deck; island and deck layout clearly visible. </O>  
    <A> Identified as U.S. Navy Nimitz-class nuclear carrier USS *Abraham Lincoln* (CVN-72), a current front-line unit. </A>  
    <P> Carrier is in port with no deployment indicators; short-term departure unlikely. Recommend routine low-frequency monitoring only. </P>
## Process Notes
    Only part of the flow is shown; after Round 3 additional rounds are omitted. The LLM should continue similar operations as needed, observing the following:
    1. Wrap reasoning in <think></think>; if tool results are returned, analyse them before thinking further or stopping.
    2. Follow the invocation format strictly; place the tool call immediately after </think>, and be meticulous with parameters.
    3. The invocation pattern can be abstracted as: analyse input -> give reason -> provide parameters.
    4. The final result must follow SOAP format exactly.
    5. Only one tool call per response; never output multiple calls at once.
    6. Use exactly the parameter names defined for the tools.
    7. When cropping, ensure the correct source image is used.
"""
\end{lstlisting}

\begin{lstlisting}[basicstyle=\ttfamily\footnotesize,breaklines=true,
    caption={Prompt for VLM}]
ROUGH_DESCRIPTION_DATASET_PROMPT="""
## Task Description:
    The current task is to briefly describe a remote-sensing image; your part of the task is:
    1. Upon receiving a satellite remote-sensing image, you must provide an overall, simple description. Make sure the description does not contain fine-grained identifying information, but category-level descriptions are allowed.
    2. Do not output any irrelevant content; our entire dialogue will be used as the dataset.
    3. Do not use code blocks and do not output acknowledgements such as "received".
    4. When writing the VLM description, do not give next-step suggestions like chat; focus only on describing the image.
    5. The task may involve multiple rounds of dialogue; strictly follow these requirements in every round.
    6. The description example is for reference only; describe according to the actual content.
## Description Example:
    This is a high-resolution satellite image showing an aircraft carrier berthed at a harbour. Numerous carrier-borne aircraft are neatly parked on the deck. A white number is present at the stern, but the resolution is insufficient to distinguish the exact digits. Roads, buildings, and vehicles are laid out regularly around the port area.
"""
\end{lstlisting}

The dataset-oriented prompt is derived from the VICoT system prompt but adapted for standalone image–text annotation. 
Compared with the SOAP-based analytic workflow, this version simplifies the LLM's final output into a concise natural-language description. 
At the end of each dialogue, the model must explicitly output the token \texttt{<end>} to mark completion, ensuring clean sample boundaries when constructing the dataset.

\subsection{MCP Tool Matching Logic}
\subsubsection{XML Format Specification and Parsing}
\begin{lstlisting}[basicstyle=\ttfamily\footnotesize,breaklines=true,caption={XML format}]
<use_mcp_tool>
  <server_name>...</server_name>
  <tool_name>...</tool_name>
  <arguments>
  { ...JSON schema... }
  </arguments>
</use_mcp_tool>
\end{lstlisting}

\begin{lstlisting}[basicstyle=\ttfamily\footnotesize,breaklines=true,
caption={GenerateToolXML workflow}]
def generate_tool_xml(tool_dict):
    for each tool in tool_dict:
        get tool name, description and input_schema
        change schema str into object
        format a xml template
        append template to xml_blocks
    append end tags
    return xml_blocks in format
\end{lstlisting}

\begin{lstlisting}[basicstyle=\ttfamily\footnotesize,breaklines=true,
caption={CheckStructuredFunctionCall workflow}]
def check_structured_function_call(LLM_output):
    regex match mcp function calling args
    return match.groups if structured_function_call else None
\end{lstlisting}

\begin{lstlisting}[basicstyle=\ttfamily\footnotesize,breaklines=true,
caption={CreateCallQuery workflow}]
def create_call_query(LLM_structured_function_calling_text):
    regex match mcp function calling args
    for each function calling match:
        call mcp server tools with args in ToolCall
        add tool call result to tool_call_list
    return tool_call_list
\end{lstlisting}

\begin{lstlisting}[basicstyle=\ttfamily\footnotesize,breaklines=true,
caption={MCPClient.call\_tools workflow}]
async def call_tools(tool_call_list): 
    for each tool_call in tool_call_list:
        find current tool's server session
        execute server function with session.call_tools
        append tool call result to result_list
    return result_list
\end{lstlisting}

\subsubsection{Environment Isolation and Runtime Decoupling}

\begin{figure}[h]
\centering
\includegraphics[width=1.0\columnwidth]{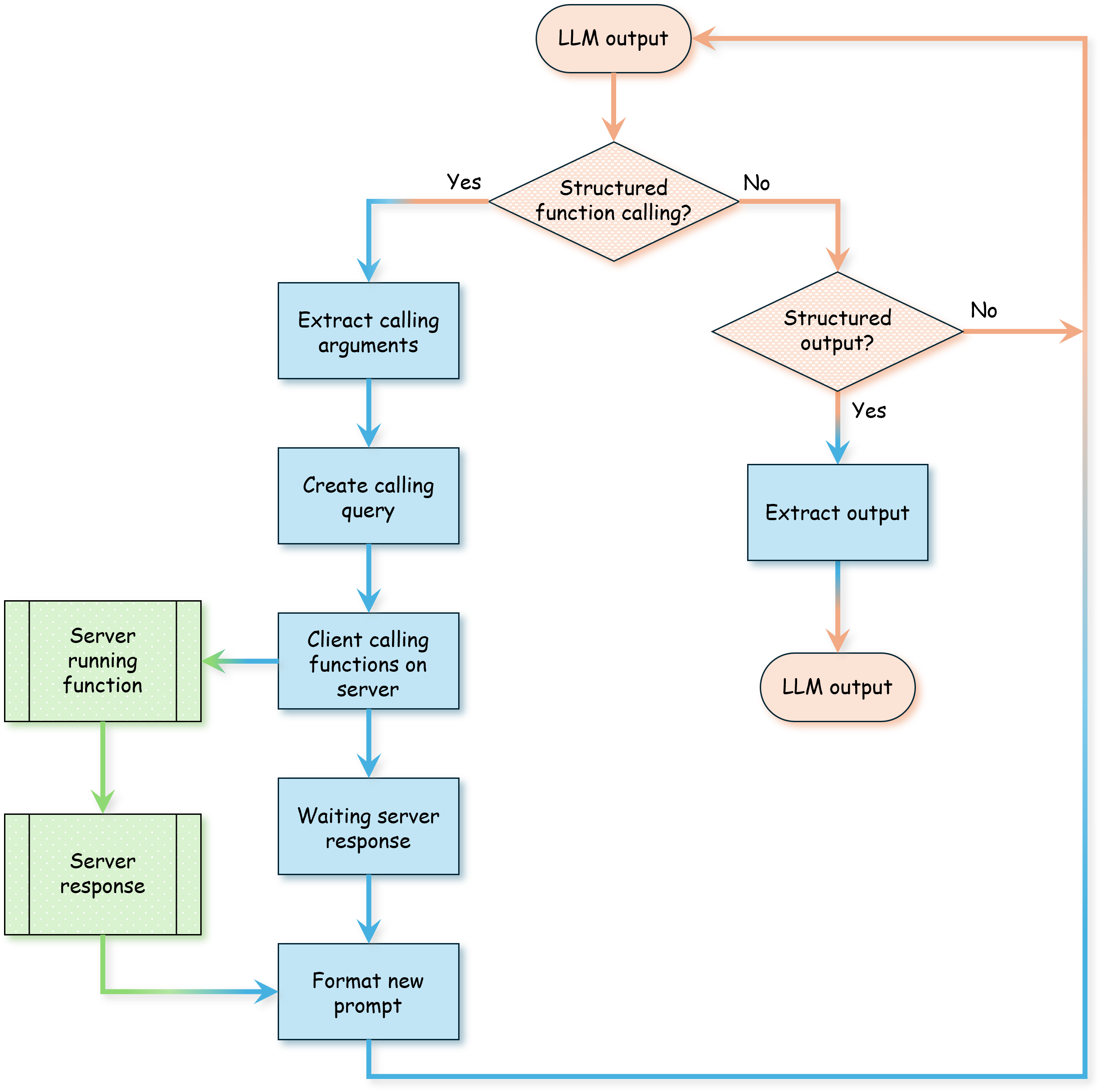} 
\caption{Isolation Agent Workflow}
\label{appendix1}
\end{figure}

Tool invocation does not rely on the execution environment of the current client. Instead, it uses the server's runtime environment directly. The server-side Python interpreter is invoked through the stdio transport protocol to handle input-output communication.When new tools are added to the server, only the server's dependencies need to be updated accordingly.

Given the simplicity of the current codebase and the requirements, more decoupled or performant remote access protocols (e.g., SSE or streamable http) are not adopted. The stdio transport protocol is sufficient for the current operational environment.

The tool list exposed in the system prompt is automatically retrieved through request to server. This architecture provides strong independence: any addition or modification of tools on the server side is automatically reflected in both the system prompt and the invocation process at runtime, without requiring any changes on the client side.

\begin{lstlisting}[basicstyle=\ttfamily\footnotesize,breaklines=true,
caption={RunVisionAgent workflow},escapeinside={(*}{*)}]
def run_vision_agent():
    input image and user prompt
    initialize connections with servers
    get server tools and store as tool_dict
    use tool_dict to format tool xml and fill in the system prompt
    get vlm rough describe response
    while not check_structured_output(response):
        start query
        get LLM response
        if check_structured_function_call(response):
            extract function calling args
            create function calling query
            find the match server and the match tool
            call server python interpreter to run function
            get server calling response
            format new prompt
        elif check_structured_output(response):
            extract structured output
            finish the task
            break
        else:
            format new prompt for another round
    clean and close session
\end{lstlisting}

\subsection{VICoT-HRSC Dataset Format}

\begin{lstlisting}[basicstyle=\ttfamily\footnotesize,breaklines=true,caption={Dataset demo},escapeinside={(*}{*)}]
[
{
"id": "image_id",
"messages": [
{
"role": "system",
"content": [
{
"type": "text",
"text": "instruction (mainly containing tool parameters)"
}
]
},
{
"role": "user",
"content": [
{
    "type": "text",
    "text": "instruction (auxiliary information)"
},
{
    "type": "image",
    "image": "./image/test.png"
}
]
},
{
"role": "assistant",
"content": [
{
    "type": "text",
    "text": "answer"
}
]
},
{
"role": "tool",
"content": [
{
    "type": "text",
    "text": "answer"
}
]
}
]
},
{
"id": "100001290",
"messages": [
{
"role": "system",
"content": [
{
    "type": "text",
    "text": "LLM system prompt (*\ref{llm_prompt}*)"

}
]
},
{
"role": "user",
"content": [
{
    "type": "text",
    "text": "The current image is a satellite remote-sensing image taken on 27 September 2023."
},
{
    "type": "image",
    "image": "100001290.png"
}
]
},
{
"role": "assistant",
"content": [
{
    "type": "text",
    "text": "This is a high-resolution satellite image showing a large vessel berthed at a quay. Multiple aircraft are neatly arranged on the deck. Surrounding areas include several berths, warehouse buildings, and shore facilities; port roads and numerous vehicles are clearly visible, giving the environment an organised layout."
}
]
},
{
"role": "user",
"content": [
{
    "type": "text",
    "text": "Please call the tools as required based on the VLM description; the image path is 100001290.png"
}
]
},
{
"role": "assistant",
"content": [
{
    "type": "text",
    "text": "<think> The description shows a large vessel berthed in a port with multiple aircraft on deck; number information may exist around the vessel, but the current description is insufficient to confirm the vessel's identity. To recognise the vessel type and hull number, I will call the GroundingDINO detection tool with prompts for vessel type, number areas, and aircraft information to support further analysis.
</think>
<use_mcp_tool>
<server_name>mcp_vision_server</server_name>
<tool_name>image_detection</tool_name>
<arguments>
{
"image_path": "100001290.png",
"txt_prompt": "military warship . cargo ship . aircraft . warship tail number . aircraft tail number . marking on tail of ship . marking of aircraft . building on port"
}
</arguments>
</use_mcp_tool>"

}
]
},
{
"role": "tool",
"content": [
{
    "type": "text",
    "text": {
"result_image_path": "100001290_annotated.png",
"boxes": [
"building on port 0.35 265 558 479 783",
"building on port 0.31 923 463 1173 828",
"military warship 0.31 186 82 547 721",
"building on port 0.28 30 85 171 179",
"white landing spot 0.27 1023 644 1116 761",
"military warship 0.26 420 687 484 787",
"military warship 0.25 489 711 541 792",
"warship tail number 0.24 488 113 523 150",
"ship island 0.24 7 547 201 703",
"building on port 0.2 1 712 67 828"
],
"vlm_response": "The image is a remote-sensing view of a harbour. After detection, ten coloured boxes with category, confidence, and coordinates are identified:

1. "building on port", conf. 0.35, (265, 558, 479, 783): a large harbour building in the central-lower part, just aft of the carrier.

2. "building on port", conf. 0.31, (923, 463, 1173, 828): a white-roof rectangular building in the lower-right; the "white landing spot" lies inside this box.

3. "military warship", conf. 0.31, (186, 82, 547, 721): the main grey aircraft carrier berthed along the quay, deck and island clearly visible.

4. "building on port", conf. 0.28, (30, 85, 171, 179): an isolated facility near greenery at the upper-left, part of the port hinterland.

5. "white landing spot", conf. 0.27, (1023, 644, 1116, 761): a long white strip, likely a helipad or unloading zone, inside box 2.

6. "military warship", conf. 0.26, (420, 687, 484, 787): a second small grey warship just below the carrier's stern.

7. "military warship", conf. 0.25, (489, 711, 541, 792): another adjacent small warship slightly to the right.

8. "warship tail number", conf. 0.24, (488, 113, 523, 150): probable hull-number area at the carrier's bow.

9. "ship island", conf. 0.24, (7, 547, 201, 703): region near water at lower-left, likely a false positive.

10. "building on port", conf. 0.20, (1, 712, 67, 828): a small structure on the waterfront at the extreme lower-left.

Overall, the carrier dominates the scene, surrounded by multiple harbour facilities and auxiliary vessels; image clarity supports multi-class military and port analysis."
}

}
]
},
{
"role": "assistant",
"content": [
{
    "type": "text",
    "text": "<think> Detection shows the main carrier (186, 82, 547, 721) and a hull-number region (488, 113, 523, 150). To enhance clarity for number recognition I will crop that region and apply binary processing.
</think>
<use_mcp_tool>
<server_name>mcp_vision_server</server_name>
<tool_name>image_binary</tool_name>
<arguments>
{
"image_path": "100001290.png",
"x1": 488,
"y1": 113,
"x2": 523,
"y2": 150
}
</arguments>
</use_mcp_tool>"
}
]
},
{
"role": "tool",
"content": [
{
    "type": "text",
    "text": {
"result_image_path": "100001290_binary_250724162738.png",
"vlm_response": "The binarised image shows a black background for water/sky and grey-to-white tones for structures.

A large grey-white vessel occupies mid-left, its deck outline sharp; orderly rows of pale aircraft sit on deck, confirming an aircraft carrier.

Left of the carrier, a quay edge and another large grey hull are partially visible.

Centre-right, two large parallel grey vessels display block-like superstructures; the rightmost shows a stepped profile, possibly support ships.

Different greyscale values separate metallic/hard surfaces (bright) from non-metal/shadow (dark). The tool successfully isolates targets from background, leaving no clutter."
}

}
]
},
{
"role": "assistant",
"content": [
{
    "type": "text",
    "text": "<think> The binarised view clarifies the carrier and deck layout, but the hull number is still not sharp. I will run super-resolution on the original to further refine the number region.
</think>
<use_mcp_tool>
<server_name>mcp_vision_server</server_name>
<tool_name>image_super_resolution</tool_name>
<arguments>
{
"image_path": "100001290.png"
}
</arguments>
</use_mcp_tool>"

}
]
},
{
"role": "tool",
"content": [
{
    "type": "text",
    "text": {
"result_image_path": "100001290_esrgan_250724162753.png",
"vlm_response": "High-resolution image of a harbour under clear skies:

1. **Carrier (left-central upper)**: large grey-blue deck ship berthed along the quay. White digits '41' are clearly painted aft. Numerous fixed-wing fighters and helicopters are neatly parked. The island with radars and a white spherical antenna sits amidships-aft.

2. **Port facilities (mid-lower left to centre)**: dense warehouses, vehicles and cranes. Multiple berths; besides the carrier, smaller craft are moored. A grid of vehicles suggests logistics or visitor parking.

3. **Upper-right berth**: a large vessel or floating platform with a wide white roof and strip structures.

4. **Upper-left**: landscaped green and orderly buildings, roads encircling, likely a visitor or park area; dozens of cars are parked.

5. **Water area**: deep-blue; two small fast boats with wakes mid-right-upper. No visible pollution.

Overall, the scene centres on a berthed carrier surrounded by port infrastructure and small craft, suitable for military and logistics analysis."
}

}
]
},
{
"role": "assistant",
"content": [
{
    "type": "text",
    "text": <think>
The super-resolved image clearly shows the white deck number '41'. Deck aircraft, island details, and dense port facilities are all distinct. Information is sufficient: carrier ID confirmed as hull number 41; the vessel appears in routine berth status. I can conclude.
</think>
The carrier with hull number "41" is berthed at the inner quay, deck crowded with aircraft; the island hosts radar, communication equipment, and a white radome. The port area is busy with warehouses, vehicles, and cranes, indicating routine logistics. Based on the number and structural features, the vessel is identified as a major fleet carrier, home-ported here. No signs of deployment readiness are present; long-range sailing is unlikely in the short term.<end>

}
]
}
]
}
]

\end{lstlisting}

\subsection{Supplementary Experimental Details}
\subsubsection{Trajectory Quality Evaluation}
\paragraph{Metrics}
Accuracy: Defined as the proportion of tool invocations in which the predicted tool name exactly matches the ground-truth tool name. An invocation is counted as correct only when the model selects the correct tool type, regardless of parameter contents. We report the overall accuracy averaged across all invocation types.

GPT-4.1 Rating (0–100): We design rating prompts to have GPT-4.1 rate each invocation’s rationality (appropriate tool selection and parameter relevance) and completeness (no missing parameters): Section \ref{gpt_rate_prompt}.
Each invocation is scored independently three times, and we report the mean.

Human Expert Score (0-100): Three remote sensing analysts with $\ge$ 10 years of experience rate each invocation's rationality and completeness using a standardized rubric. Inter-rater reliability is first assessed (Cohen's $\kappa > 0.8$), after which we report the averaged score.

BLEU (0–1): We compute the BLEU-4 score between the serialized invocation sequence and the reference using NLTK, including 1–4 grams, and normalize the score to [0,1].

\paragraph{Implementation Details}
Parameters: temperature=0, max output tokens=2048;
MCP tool suite integrated via a standard XML interface; 
Each method is evaluated with three random seeds; we report the mean and standard deviation for each metric. 
Hardware Platform: NVIDIA RTX 4090 GPU (24GB VRAM), CUDA 11.8.

\subsubsection{GPT-4.1 rating prompts}
\label{gpt_rate_prompt}
\begin{lstlisting}[
basicstyle=\ttfamily\footnotesize,
breaklines=true,
caption={VICoT Evaluation Instruction Prompt}
]
---Role---
You are an expert evaluator specializing in multimodal tool-augmented reasoning,
remote sensing image understanding, and structured XML-based tool invocation.
Your responsibility is to evaluate the quality of VICoT agent reasoning trajectories
generated under different configurations or sampling strategies.

---Goal---
You will rate a single reasoning trajectory sampled from VICoT's multi-turn
vision-language-tool pipeline. Evaluate the trajectory using the four criteria below:

- Tool Correctness:
  Whether tool invocations follow the specification strictly:
  correct tool name, valid parameter formats, reasonable parameter values,
  and proper XML structure.

- Visual Groundedness:
  Whether each reasoning step and invoked tool are grounded in the image evidence,
  and whether VLM textual descriptions align with visual regions.

- Information Gain:
  Whether each step extracts new, useful, and non-redundant information
  from tools or VLM feedback, progressively reducing uncertainty.

- Reasoning Stability:
  Whether the reasoning flow is stable, coherent, and consistent across turns,
  without hallucinated tool calls or unnecessary loops.

For each trajectory, provide a continuous score for all four criteria following
the rubric above, and then compute and output the final total score in the range [0, 100].
\end{lstlisting}

\begin{lstlisting}[
basicstyle=\ttfamily\footnotesize,
breaklines=true,
caption={VICoT Evaluation Input Prompt}
]
Here is the reasoning trajectory produced by the VICoT agent.

You are provided with:
- The original image description: {Image_Description}
- The full multi-turn reasoning trajectory (LLM thoughts, tool invocations, and tool results): 
  {Full_Trajectory}
- The final answer produced by the agent: {Final_Output}

Please evaluate this trajectory according to the four criteria defined in the Instruction Prompt
(Tool Correctness, Visual Groundedness, Information Gain, Reasoning Stability).

Assign a continuous score for each criterion, and compute a total score (0-100).

Output your evaluation strictly in the JSON format below:

{
  "Tool_Correctness": {
    "Score": *float in [0,40]*,
    "Explanation": "reasoning"
  },
  "Visual_Groundedness": {
    "Score": *float in [0,25]*,
    "Explanation": "reasoning"
  },
  "Information_Gain": {
    "Score": *float in [0,20]*,
    "Explanation": "reasoning"
  },
  "Reasoning_Stability": {
    "Score": *float in [0,15]*,
    "Explanation": "reasoning"
  },
  "Total_Score": *float in [0,100]*
}
\end{lstlisting}

\subsection{Error cases}
Error cases arise primarily from the inherent uncertainty in LLM outputs.
Even with temperature set to 0, the model may still generate incomplete or incorrect tool-call arguments, leading to mismatches in file paths, bounding-box coordinates, or parameter types. Because the LLM cannot reliably guarantee the correctness of every invocation, the agent depends on MCP’s structured error feedback (e.g., file-not-found, invalid region, type errors) to identify failure modes. These error messages are then fed back into the reasoning loop, allowing the agent to adjust its next tool invocation accordingly.
\begin{lstlisting}[basicstyle=\ttfamily\footnotesize,breaklines=true,
caption={Parameters error}]
{
  "_meta": null,
  "content": [
    {
      "type": "text",
      "text": "Error: OpenCV(4.11.0) /io/opencv/modules/imgcodecs/src/loadsave.cpp:929: error: (-215:Assertion failed) !_img.empty() in function 'imwrite'\n",
      "annotations": null
    }
  ],
  "isError": true
},
{
  "_meta": null,
  "content": [
    {
      "type": "text",
      "text": "Error: [Errno 2] No such file or directory: 'datasets/label/temp/temp/100001035_cropped_250807174005_esrgan_250807174032_boxes.txt'",
      "annotations": null
    }
  ],
  "isError": true
}


\end{lstlisting}

\begin{lstlisting}[basicstyle=\ttfamily\footnotesize, breaklines=true, caption={Tool usage error}]
{
    "role": "assistant",
    "content": [
        {
            "type": "text",
            "text": "<think> The cropping operation encountered an error, likely because the specified region exceeded the image boundaries or the image path was invalid. I will first check the dimensions of the grayscale image to confirm whether the cropping region is valid and adjust the parameters to avoid out-of-bound errors.
</think>
<use_mcp_tool>
    <server_name>mcp_vision_server</server_name>
    <tool_name>image_detection</tool_name>
    <arguments>
    {
        "image_path": "datasets/label/temp/100001430_cropped_250807191335_gray_250807191356.png",
        "txt_prompt": "image size"
    }
    </arguments>
</use_mcp_tool>"
        }
    ]
}
\end{lstlisting}

\end{document}